\documentclass[preprint,12pt]{elsarticle}

\usepackage{amssymb}
\usepackage{amsmath}
\usepackage{graphicx}%
\usepackage{multirow}%
\usepackage{amsmath,amssymb,amsfonts}%
\usepackage{amsthm}%
\usepackage{mathrsfs}%
\usepackage[title]{appendix}%
\usepackage{xcolor}%
\usepackage{textcomp}%
\usepackage{manyfoot}%
\usepackage{booktabs}%
\usepackage{algorithm}%
\usepackage{algorithmicx}%
\usepackage{algpseudocode}%
\usepackage{listings}%
\usepackage{pifont}%
\usepackage{hyperref}%

\journal{}

\begin{document}

\begin{frontmatter}

\title{PointViG: A Lightweight GNN-based Model for Efficient Point Cloud Analysis}

\author[label1]{Qiang Zheng}
\author[label2]{Yafei Qi}
\author[label3]{Chen Wang}
\author[label1]{Chao Zhang}
\author[label1]{Jian Sun \corref{cor1}}

%% Abstract
\begin{abstract}
In the domain of point cloud analysis, despite the significant capabilities of Graph Neural Networks (GNNs) in managing complex 3D datasets, existing approaches encounter challenges like high computational costs and scalability issues with extensive scenarios. These limitations restrict the practical deployment of GNNs, notably in resource-constrained environments. To address these issues, this study introduce \textbf{Point} \textbf{Vi}sion \textbf{G}NN (PointViG), an efficient framework for point cloud analysis. PointViG incorporates a lightweight graph convolutional module to efficiently aggregate local features and mitigate over-smoothing. For large-scale point cloud scenes, we propose an adaptive dilated graph convolution technique that searches for sparse neighboring nodes within a dilated neighborhood based on semantic correlation, thereby expanding the receptive field and ensuring computational efficiency. Experiments demonstrate that PointViG achieves performance comparable to state-of-the-art models while balancing performance and complexity. On the ModelNet40 classification task, PointViG achieved 94.3\% accuracy with 1.5M parameters. For the S3DIS segmentation task, it achieved an mIoU of 71.7\% with 5.3M parameters. These results underscore the potential and efficiency of PointViG in point cloud analysis.
\end{abstract}


% \begin{highlights}
% \end{highlights}

%% Keywords

\end{frontmatter}

\section{Introduction}

\begin{figure}[ht]
    \centering
    \includegraphics[width=0.7\linewidth]{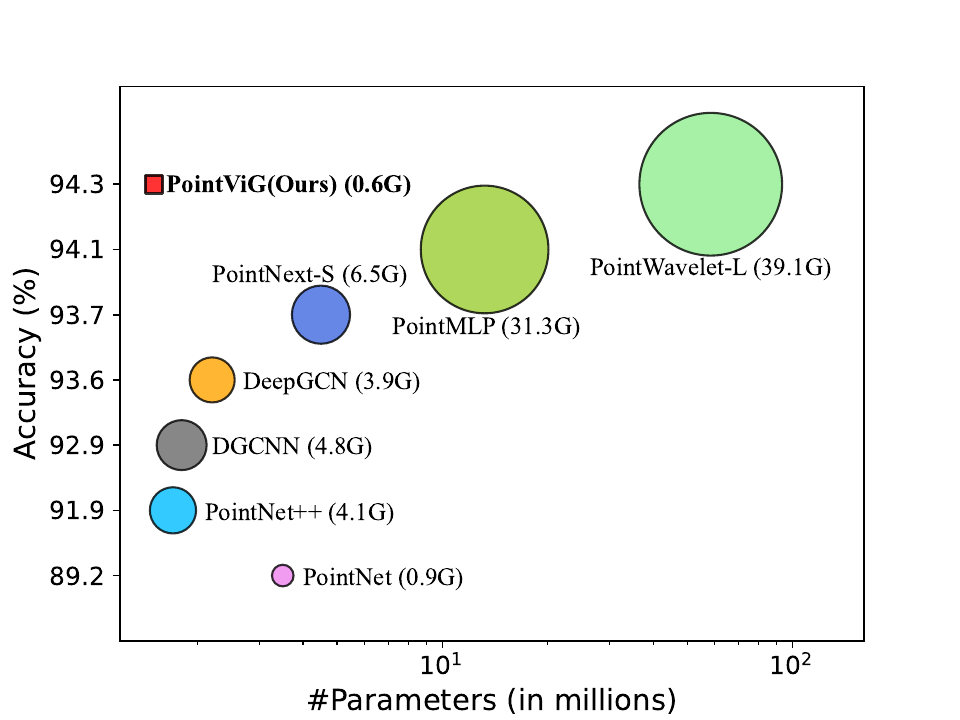}
    \caption{Comparison of classification accuracy among representative models, where bubble areas correspond to the number of floating-point operations (FLOPs). The specific FLOPs values (in billions) are provided in textual notation. PointViG surpasses other models, demonstrating superior performance with fewer parameters and FLOPs. This highlights PointViG's optimal trade-off between performance and complexity. }
    \label{complexity}
\end{figure}

In recent years, the advancement of 3D vision technology in domains such as robotics, autonomous driving, and 3D reconstruction has propelled point cloud analysis into the forefront of 3D understanding. This has garnered extensive attention from both academia and industry. In contrast to the structured pixel representation, a point cloud is an assemblage of unordered and dispersed points, introducing challenges such as disorder, irregularity, sparsity, and noise. These inherent characteristics render point cloud analysis a challenging endeavor. 

Previous endeavors have employed techniques such as voxelization~\cite{VoxNet2015, SPLATNet2018, OctNet2017}, or projection~\cite{2015Multi, multi2021}. However, the processes of voxelization or projection come at the cost of losing fine-grained or structured information. PointNet~\cite{PointNet2017} stands out as the pioneering point-based method. Subsequent to its introduction, numerous methods have emerged, adopting the raw point clouds directly as inputs. Categorically, based on the local feature aggregation, point-based methods fall into distinct groups, including MLP-based~\cite{PointNetplus2017}, Convolutional Neural Network (CNN)-based~\cite{PointConv2019}, Graph Neural Network (GNN)-based~\cite{DGCNN2019}, and Transformer-based methods~\cite{PT2021}. Among these, graph-based approaches treat scattered point clouds as graph data, and GNNs have emerged as a promising solution for non-Euclidean data, such as point clouds. 

GNNs are highly effective for point cloud analysis but come with significant computational costs. This complexity mainly stems from the extensive matrix operations and iterative message-passing mechanisms employed in high-dimensional feature spaces. These operations scale quadratically with the graph's node count and are heavily influenced by feature dimensionality. Additionally, the large scale of neighbor node tensors, which often exceed the size of the graph nodes themselves, further increases the computational burden. For example, DGCNN~\cite{DGCNN2019} requires re-searching neighbor nodes based on updated features before each graph convolution without down-sampling, resulting in substantial computation costs. It also fails to effectively control the over-smoothing phenomenon, limiting potential performance improvements. MRGCN~\cite{DeepGCNs2021} attempts to enhance performance by constructing deeper GNN networks, but this approach significantly increases time and space complexity, while offering relatively limited performance gains.

Moreover, these limitations are exacerbated in large-scale scenarios, making the direct application of GNNs for analyzing large scenes impractical. To address this issue, DGCNN~\cite{DGCNN2019} and PointWavelet~\cite{PointWavelet2023} divide the scene into multiple blocks, while SPG~\cite{SPG2018} divides the scene into homogeneous patches through unsupervised graph partitioning before semantic segmentation. Although these methods alleviate computational pressure to some extent, they also limit precise semantic segmentation capabilities. AdaptConv~\cite{Adaptive2021} combines local feature extraction with a pyramid structure to better preserve connections between nodes. Despite progress in expanding the receptive field, there remains an irreconcilable contradiction between expanding the receptive field and controlling computational cost.

This paper introduces a GNN-based framework that is computationally efficient and widely applicable to various point cloud analysis tasks. At the core of this framework is a lightweight, plug-and-play graph convolution module. Traditional graph convolution operations typically involve processing two tensors: one related to the node itself and the other to its neighboring nodes. The scale of the neighbor node tensor is often much larger than that of the node itself. The proposed module optimizes this computational process by effectively compressing operations on the neighbor node tensor, thereby significantly reducing the overall computational load. Furthermore, to address the pervasive issue of over-smoothing in GNNs, which affects the diversity and representational power of node features, the module incorporates strategies specifically designed to enhance feature diversity, effectively mitigating this problem.

Additionally, this paper proposes an adaptive dilated graph convolution strategy for efficiently handling large-scale scenarios. This strategy first searches for a dilated subgraph for each node in low-dimensional geometric space, and then adaptively searches for sparse neighboring nodes within the subgraph based on high-dimensional semantic associations. Unlike traditional dilated convolutions, this method dynamically determines sampling points based on semantic associations, effectively expanding the receptive field without incurring excessive computational costs and ensuring that key semantic information is preserved during the sampling process.

Overall, the GNN-based point cloud analysis framework proposed in this paper, through its lightweight graph convolutional module and adaptive dilated graph convolution strategy, not only improves computational efficiency but also enhances processing capabilities for large point clouds, providing an efficient solution for point cloud analysis. The primary contributions can be summarized as follows:
\begin{itemize}
\item   We introduce an effective GNN-based network PointViG designed specifically for the analysis of point clouds.

\item   We present an efficient graph convolutional module. This module substantially reduces the computational complexity of the network while effectively mitigating the over-smoothing phenomenon.

\item   We propose an adaptive dilated graph convolution strategy to overcome the computational challenges associated with large point cloud scenes. This strategy expands the receptive field without imposing excessive computational overhead, thus enhancing scalability for larger scenes.

\item   We validate the performance of PointViG through extensive experiments across various tasks. The results demonstrate PointViG's competitiveness with state-of-the-art (SOTA) approaches. Notably, PointViG achieves comparable performance with lower complexity, achieving an optimal balance between performance and computational cost.
\end{itemize}

\section{Related Works}	
\subsection{Graph Neural Networks}
Graph Neural Networks (GNNs) were initially proposed in the seminal works~\cite{GNN2005, GNN2009}. GNNs can be categorized into spatial-based and spectral-based models, depending on the utilized operators. The framework introduced in \cite{NN4G2009} pioneered an early spatial-based Graph Convolutional Network (GCN) by integrating non-recursive layers. Notably, recent advancements have witnessed the introduction of various spatial-based GCN variants, exemplified by \cite{Diffusion2016, Message2017, Learning2016}. In contrast, the spectral-based GCN was initially introduced by \cite{Spectral2013}, drawing inspiration from spectral graph theory. Subsequently, numerous spectral-based GCN models, such as \cite{Fast2016, Clustering2020, Semi2016} have emerged. GCNs find widespread application in handling non-Euclidean data across real-world domains, including social networks, biochemical graphs, and citation networks. In computer vision, GCNs play a pivotal role in diverse applications, such as action recognition, scene graph generation, and point cloud analysis. 
Numerous studies such as \cite{Structural2016, Temporal2018} construct graph data by establishing connections among human joints, subsequently employing Graph Convolutional Networks (GCNs) for the recognition of human actions. Scene graph generation, a process that automatically generates semantic graph structures representing objects and their relationships based on images, integrates object detection with GCN, as illustrated by \cite{Scene2017, Generation2020}. GCNs inherently lend themselves to the processing of unstructured data like point clouds, enabling tasks such as point cloud classification and segmentation. This corresponds to graph classification and node classification, exemplified by \cite{SPG2018, DGCNN2019}. Notably, Approach~\cite{ViG2022} introduces a novel approach, splitting regular images into patches and treating them as nodes. Leveraging GCN for processing these nodes yields promising performance, showcasing the adaptability and efficacy of GCNs in various image processing tasks.

\subsection{Point-based Methods}
To address irregularities in point clouds, contemporary models adopt a direct input of the raw point cloud. Research endeavors, predicated on local feature extraction methods, are categorized into MLP-based approaches such as ~\cite{PointNet2017, PointNetplus2017, PointMLP2022, PointNext2022, 2023JSNet++}, CNN-based methods exemplified by ~\cite{PointCNN2018, PAConv2021, SpiderCNN2018, KPConv2019}, attention-based techniques including~\cite{Set2019, Point2Sequence2019, Attentional2018, Modeling2019, Fast2022, PyramidPC2021, PointM2AE2022, PointBERT2022, Mask2023, 2023GTNet, 2023I2PMAE, 2023AFGCN, 2023DCNet, 2023LCPFormer}, and graph-based approaches. In the graph-based paradigm, points are treated as graph nodes, and edges are established based on spatial and feature relationships, aligning with the natural representation of point clouds as graph data. The inception of GCN for point cloud dates back to ~\cite{Semi2016}, with subsequent works by exploring local features from neighboring points~\cite{DGCNN2019, Spherical2021, Pointwise2018, Mining2018, Hierarchical2019}. DGCNN~\cite{DGCNN2019} dynamically constructs a graph in the updated feature space, introducing the EdgeConv operator to aggregate features. 3D-GCN~\cite{Deformable2020} enhances local feature extraction through a learnable kernel, ensuring shift and scale-invariance. Approach~\cite{Adaptive2021} introduces AdaptConv, generating convolution kernels adaptively based on point features to capture diverse relationships. Point2Node~\cite{Point2Node2019} dynamically integrates relationships between nodes, employing a gating mechanism for adaptive feature aggregation at the channel level. DeepGCN~\cite{DeepGCNs2021} incorporates concepts such as residual connectivity and dilated convolution to train very deep GNN. Method~\cite{Spectral2018} utilizes spectral graph convolution for feature extraction and proposing a recursive pooling operation based on spectral clustering partitioning. Some approaches~\cite{SPG2018, Oversegmentation2019} partition the scene into patches, treating them as super-points, and subsequently perform semantic prediction on each super-node. 

\section{Method}
In this section, we commence with an introduction to the proposed PointViG Module, in Sec. \ref{Module}. Subsequently, Sec. \ref{Adapt-dilated} introduces the adaptive dilated graph convolution, specifically designed for use in large point cloud scenarios. Finally, Sec. \ref{Architecture} provides detailed insights into the network architecture.

\subsection{PointViG Module} \label{Module}
\begin{figure}[ht]
    \centering
    \includegraphics[width=\linewidth]{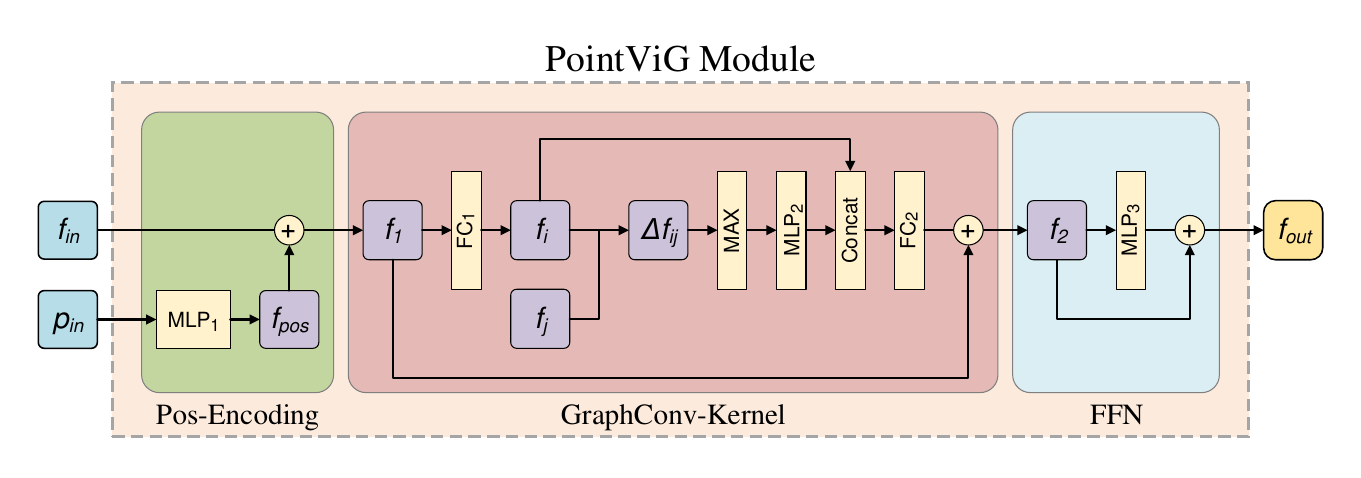}
    \caption{PointViG Module acts as the basic module in PointViG framework.}
    \label{PV-module}
\end{figure}

As depicted in Fig. \ref{PV-module}, the PointViG Module is primarily composed of two main components: the graph convolutional kernel (referred to as "GraphConv Kernel" in Fig. \ref{PV-module}) and several supplementary components (referred to as "Pos-Encoding" and "FFN" in Fig. \ref{PV-module}). In this section, we initially elucidate the mathematical notation involved in graph convolution operations, followed by a detailed explanation of the working mechanism of the GraphConv Kernel, and conclude with an introduction to the supplementary components.

\subsubsection{Mathematical Notation}
Given an input point cloud with $N$ points, where each point is represented by a position vector $p_{i} \in \mathbb{R}^{3}$ and a corresponding feature vector $f_{i} \in \mathbb{R}^{D}$, with $D$ denoting the feature dimension and $i = 1, 2, \ldots, N$, we treat these features as nodes within a graph denoted as $\mathcal{V} = \{v_{1}, v_{2}, \ldots, v_{N}\}$. For each node, we identify neighbors $\mathcal{N}(v_{i})$ in the feature space and establish edges $e_{ij}$ connecting $v_{i}$ to its neighbors $v_{j}$. Consequently, the point cloud is transformed into a graph $\mathcal{G} = (\mathcal{V}, \mathcal{E})$, where $\mathcal{E}$ represents all edges.

\subsubsection{GraphConv-Kernel}
The GraphConv Kernel is tasked with performing the fundamental graph convolution operations. It comprises two fully connected layers, FC$_{1}$ and FC$_{2}$, positioned at the input and output, respectively. Considering the output feature of the FC$_{1}$ layer, let $f_{i}$ represent the central node feature, and $f_{j}\in\mathcal{N}(f_{i})$ denote a set of neighborhood features. To capture local characteristics in the feature domain, we compute the difference $\Delta f_{ij} = f_{j} - f_{i}$ for each neighborhood feature relative to the central node. The resulting vectors $\Delta f_{ij}$ undergo max-pooling and are then input to a multilayer perceptron (MLP). Each MLP layer is followed by batch normalization and nonlinear activation to facilitate nonlinear transformation. Finally, the transformed $\Delta f_{ij}$ and the central feature $f_{i}$ are concatenated to integrate information from central and neighboring nodes. This concatenated feature is further processed by FC$_{2}$.

GraphConv Kernel adopts an efficient design that ensures effective graph neighborhood feature aggregation. The GraphConv Kernel integrates global information denoted by the central node $f_{i}$ with local information represented by $\Delta f_{ij}$ in a high-dimensional space, thereby enhancing feature characterization. Additionally, the design avoids premature involvement of the central node during the transformation of local feature $\Delta f_{ij}$ by $\text{MLP}_2$, preserving the independence of these two features, contrary to DGCNN~\cite{DGCNN2019}. Moreover, the tensors of neighboring nodes often exceed the scale of the graph itself. In GraphConv Kernel, operations related to these neighbor node tensors are simplified to subtraction and pooling. By applying the majority of operations within the GraphConv Kernel and the entire PointViG Module directly to the nodes themselves and compressing computations related to neighbors, the computational burden on the network has been significantly reduced.

The operational mechanism of the GraphConv Kernel is illustrated in the following formula, where the notation within the formula is consistent with that of the preceding text.

    \begin{equation}
        \label{fc1}
        \it{f_{i}}=\rm{\bf{FC_{1}}}(\it{f_{\rm1}})
    \end{equation}	 
    \begin{equation}
        \label{kernel}
        \it{f_{max}}=\rm{\bf{MAX}}\{\it{f_{j}-f_{i}}|\it{f_{j}\in \mathcal{N}(f_{i}) \in G}\}    
    \end{equation}	
    \begin{equation}
        \label{fc2}
        \it{f_{\rm2}}=\rm{\bf{FC_{2}}}\{\rm{\bf{Concat}}[\rm{\bf{MLP_{2}}}(\it{f_{max}}),\it{f_{i}}]\}+\it{f_{\rm1}}
    \end{equation}

\subsubsection{Supplementary Components}
The PointViG Module, which includes the aforementioned GraphConv kernel along with the Pos-Encoding (Position Encoding) and the Feed-Forward Network (FFN), significantly enhances feature diversity. For a given input node $p_{\text{in}}$ with its corresponding feature $f_{\text{in}}$, the Pos-Encoding block utilizes a 3-layer MLP designed to encode positional information. The resulting position encoding feature is denoted as $f_{\text{pos}}$ and is merged with the input feature $f_{\text{in}}$. The position encoding serves as a supplement to the input information of graph convolution, akin to edge features, providing an effective supplement to the input information of graph convolution. Subsequently, the updated feature $f_{1}$ is passed to the GraphConv Kernel and the FFN block for graph convolution and subsequent non-linear transformation. Alongside the FFN block, the Pos-Encoding block assists in mapping the inputs and outputs of the GraphConv kernel to a different domain, playing a crucial role in enhancing feature diversity and mitigating the over-smoothing phenomenon. The statistical results regarding feature diversity validate this assertion. The overall sequence of operations within the PointViG Module is summarized as follows:

    \begin{equation}
        \it{f_{\rm1}}=\it{f_{\rm in}}+\rm{\bf{MLP_{1}}}(\it{p_{in}})
    \end{equation}
    \begin{equation}
        \label{graphconv-kernel}
        \it{f_{\rm2}}=\rm{\bf{GraphConv}}(\it{f_{\rm1}})
    \end{equation}
    \begin{equation}
        \it{f_{\rm out}}=\rm{\bf{MLP_{3}}}(\it{f_{\rm2}})+\it{f_{\rm2}}
    \end{equation}

\subsection{Adaptive Dilated Graph Convolution} \label{Adapt-dilated}
To enhance semantic perception without significantly increasing computational demand, we introduce the adaptive dilated graph convolution. The computational complexity of distance computation is $\mathcal{O}(N^{2}d)$ when seeking neighbors for all nodes in the graph, where $N$ and $d$ represent the number of nodes and the feature dimension, respectively. Consequently, for scenarios involving large point clouds, direct querying of neighboring nodes across the entire graph becomes impractical. Notably, in expansive scenes with multiple targets, there exists a substantial correlation in semantics between points within an individual target and its nearby region. Conversely, the semantic association between objects situated at a considerable distance is typically weak. We empirically assume that semantic associations exhibit non-uniformity within extensive scenes. Based on this empirical assumption, this study divides the neighbor node search in large point cloud scenes into two stages. Initially, in the low-dimensional geometric space, spatial neighbor points for each point are identified to form its subgraph; subsequently, within the subgraph, sparse neighbor nodes are adaptively searched based on the correlations in the high-dimensional semantic space. Owing to the adaptive sparse sampling strategy, it becomes feasible to define a broader subgraph, consequently extending the perceptual field of each node significantly.

The adaptive dilated graph convolution expands the receptive field while concurrently minimizing computational overhead. Analyzing the complexity, let $N$ represent the scene's total node count, $d$ the feature dimension with the condition that $d \gg 3$, and $m$ the number of nodes within each subgraph, where $m \ll N$. The proposed sampling strategy is divided into a two-step process. Initially, subgraphs are formed based on Euclidean distances, which introduces a complexity of $\mathcal{O}(3N^2)$. Subsequently, the process involves identifying neighbors for a central node within each subgraph among $m$ nodes, yielding a complexity of $\mathcal{O}(md)$ per subgraph. When aggregated across $N$ subgraphs, this results in a complexity of $\mathcal{O}(mdN)$. Thus, the overall computational complexity is encapsulated by $\mathcal{O}(3N^2 + mdN)$. The ratio $\alpha$, which compares the complexities of the proposed method to the direct sampling approach, is expressed as follows: 
\begin{equation}
\alpha = \frac{\mathcal{O}(3N^{2}+mdN)}{\mathcal{O}(N^{2}d)} = (\frac{3}{d} + \frac{m}{N}) \ll 1
\end{equation}

The distinction between the adaptive dilated graph convolution and the conventional dilated convolution applied in CNN lies in the fact that the conventional dilated convolution typically specifies sparse neighbors with pre-established sampling rules within a local patch based on spatial relationships. In contrast, the adaptive dilated graph convolution dynamically selects neighborhood nodes within the subgraph adaptively, based on the strength of semantic associations. 

In experimental settings, where samples from ModelNet40 and ShapeNet denote individual objects characterized by strong semantic associations among points, we choose to directly query neighboring nodes throughout the entire graph. Conversely, for expansive S3DIS scenes with multiple targets, we employ the adaptive dilated graph convolution to efficiently search for neighborhood nodes.

\subsection{Network Architecture}\label{Architecture}
\begin{figure*}[htbp]
    \centering
    \includegraphics[width=0.99\linewidth]{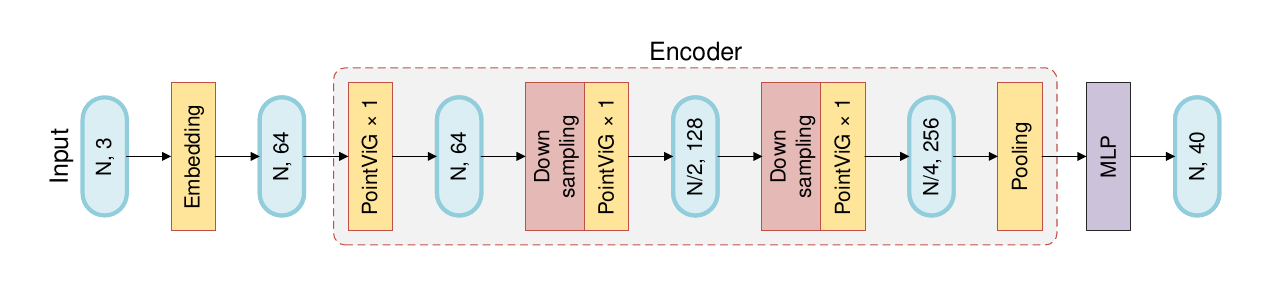}
    \caption{The PointViG architecture designed for classification.}
    \label{Cls_network}
\end{figure*}

\begin{figure*}[htbp]
    \centering
    \includegraphics[width=0.99\linewidth]{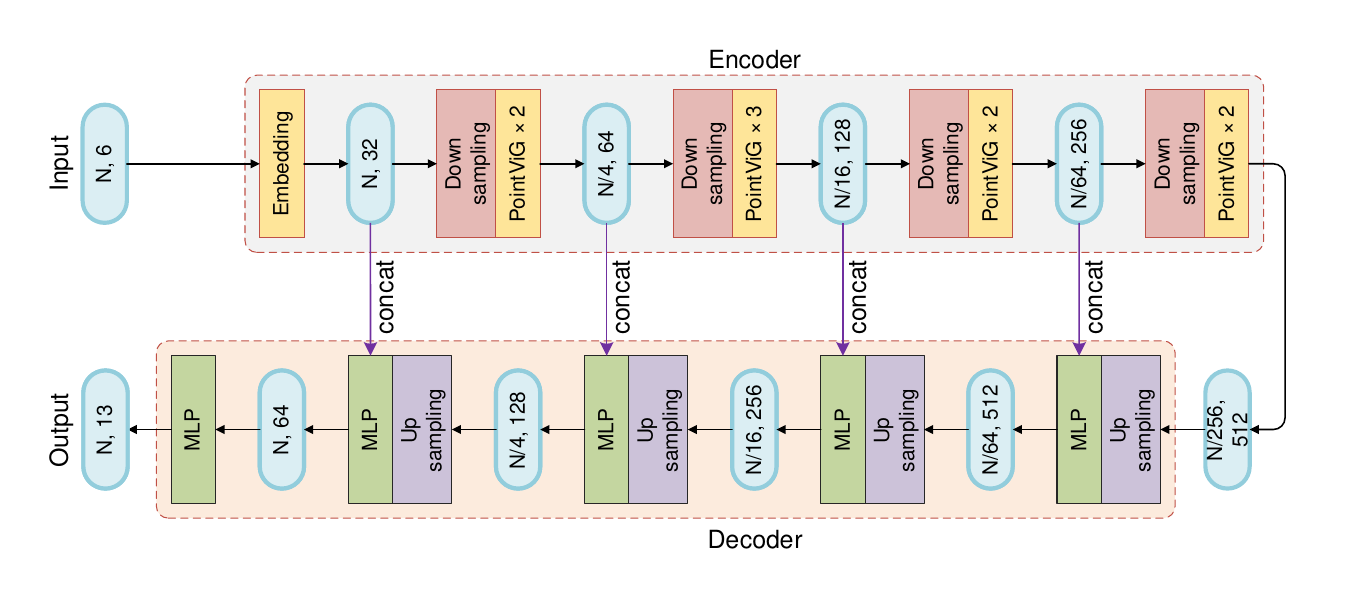}
    \caption{The PointViG architecture designed for semantic segmentation.}
    \label{Seg_network}
\end{figure*}

The PointViG architecture, designed for classification and semantic segmentation, is plotted in Fig. \ref{Cls_network} and Fig. \ref{Seg_network}, respectively. We devised a pyramid encoder backbone for progressive feature extraction. For classification, each encoder stage globally searches for neighborhood nodes, and the resulting output undergoes mean-pooling before being fed to an MLP for classification predictions. In the segmentation task, an asymmetric structure is adopted between the encoder and decoder. Each decoder stage integrates an upsampling layer and an MLP, facilitating the continuous mapping of features from the sparse point set to the dense set. The output of each encoder stage is concatenated with the corresponding decoder stage's upsampling layer output via a skip connection. All experiments were conducted using PyTorch~\cite{Pytorch2019} on a TITAN XP GPU.

\section{Experiments}
In this section, we initially assess the model's performance across tasks including classification, part segmentation, and scene semantic segmentation. Subsequently, we conduct comprehensive experiments to validate the network design, and showcase some visualization results.

\subsection{Classification}
The network undergoes evaluation on the ModelNet40 classification task, consisting of 12,311 CAD models distributed among 40 classes. The training set encompasses 9,843 models, while the remaining 2,468 are designated for testing. In accordance with the methodology presented in PointNet~\cite{PointNet2017}, 1,024 point clouds are uniformly sampled from the models, with their corresponding $(x, y, z)$ coordinates stored as input. Data augmentation is applied through a random scale transformation with parameters $[0.7, 1.0/0.7]$. For the classification task, the encoder comprises three stages, each housing a single block. The input channels for the three stages are 64, 128, and 256, with downsampling ratios of 1, 2, and 2, respectively. A batch size of 32 is employed. The Adam optimizer is chosen, with an annealing learning rate with a 25-epoch period. Within each cycle, the learning rate progressively decreases from 0.001 to $1.0 \times 10^{-5}$.

The classification results are presented in Tab. \ref{tab-cls}, with evaluation metrics including mean class accuracy (mAcc) and overall accuracy (OA). PointViG demonstrates superior performance compared to other models, even without employing auxiliary inputs and the voting strategy.

\begin{table}[ht]
    \centering
    \footnotesize
    \setlength{\tabcolsep}{2.0mm}
    \begin{tabular}{l|cccc}
    % \begin{tabularx}{\textwidth}{l|cccc}
        \toprule[1pt]
        \textbf{Method} & Input & \#points & mAcc(\%) & OA(\%) \\
        \midrule[0.3pt]
        % \midrule[0.3pt]
        PointNet~\cite{PointNet2017}					& xyz           & 1k    & 86.0  & 89.2 \\
        PointNet++(MSG)~\cite{PointNetplus2017}			& xyz, nor      & 5k    & -     & 91.9 \\
        SpecGCN~\cite{Spectral2018}					    & xyz           & 1k    & -     & 92.1 \\
        3D-GCN~\cite{Deformable2020}				    & xyz           & 1k    & -     & 92.1 \\
        PointCNN~\cite{PointCNN2018}					& xyz           & 1k    & 88.1  & 92.2 \\
        PCNN~\cite{PCNN2018}                            & xyz           & 1k    & -     & 92.3 \\
        SpiderCNN~\cite{SpiderCNN2018}				    & xyz, nor      & 5k    & -     & 92.4 \\
        CSANet~\cite{2022CSANet}                        & xyz           & 1k    & 89.9  & 92.8 \\
        DGCNN~\cite{DGCNN2019}					        & xyz           & 1k    & 90.2  & 92.9 \\
        RS-CNN~\cite{RSCNN2019} w/o vot.                & xyz           & 1k    & -     & 92.9 \\
        KPConv~\cite{KPConv2019}					    & xyz           & 6.8k  & -     & 92.9 \\
        Point2Node~\cite{Point2Node2019}                & xyz           & 1k    & -     & 93.0 \\
        PCT~\cite{PCT2021} w/o vot.                     & xyz           & 1k    & -     & 93.2 \\
        PointNext~\cite{PointNext2022}                  & xyz           & 1k    & 90.8  & 93.2 \\
        DTO-Net~\cite{2022DTONet}                       & xyz           & 1k    & 91.4  & 93.3 \\
        SO-Net~\cite{SO-Net2018}						& xyz, nor      & 5k    & -     & 93.4 \\
        AdaptConv~\cite{Adaptive2021}			        & xyz           & 1k    & 90.7  & 93.4 \\
        PointConT~\cite{2024PointConT}                  & xyz           & 1k    & -     & 93.5 \\
        DeepGCN~\cite{DeepGCNs2021}                     & xyz           & 1k    & 90.9  & 93.6 \\
        PointMixer~\cite{pointmixer2021}                & xyz           & 1k    & 91.4  & 93.6 \\
        PT~\cite{point-trans2021}                       & xyz           & 1k    & 90.6  & 93.7 \\
        CurveNet~\cite{CurveNet2021}                    & xyz           & 1k    & -     & 93.8 \\
        PointMLP~\cite{PointMLP2022}                    & xyz           & 1k    & 90.9  & 94.1 \\
        \midrule[0.3pt]
        PointViG (Ours)                                  & xyz           & 1k    & 91.2  & 94.3 \\
        \bottomrule[1pt]
    \end{tabular}
    \vspace{5pt}
    \caption{Classification results for the ModelNet40 dataset.}
    \label{tab-cls}
\end{table}

\subsection{Part Segmentation}
PointViG undergoes additional evaluation using the ShapeNetPart dataset to address the part segmentation task. This dataset comprises 16,880 samples across 16 classes. Each object encompasses between 2 to 6 parts, resulting in a dataset total of 50 parts. The ShapeNet experiments in this paper utilize only the coordinates as input. The network adopts an encoding-decoding asymmetric structure. The encoder comprises 3 stages with downsampling ratios of 1, 4, and 4. The neighborhood size for each stage is set to 32, and the model undergoes training for 100 epochs. The reported results, encompassing mean class IoU (mIoU) and mean instance IoU (IoU), are detailed in Tab. \ref{tab-partseg}. PointViG exhibits comparable performance to advanced methods.

\begin{table}[ht]
    \centering
    \footnotesize
    %\scriptsize
    \setlength{\tabcolsep}{3.5mm}
    \begin{tabular}{l|cc} 
        \toprule[1pt]
        \textbf{Method}                & mIoU(\%) & IoU(\%)  \\
        \midrule[0.3pt]
        PointNet~\cite{PointNet2017}   & 80.4 & 83.7 \\
        SO-Net~\cite{SO-Net2018}       & 81.0 & 84.9 \\
        PointNet++~\cite{PointNetplus2017} & 81.9 & 85.1 \\
        3D-GCN~\cite{Deformable2020}   & 82.1 & 85.1 \\
        DGCNN~\cite{DGCNN2019}         & 82.3 & 85.2 \\
        PCNN~\cite{PCNN2018}           & -    & 85.1 \\
        PCNN~\cite{PCNN2018}           & -    & 85.1 \\
        PointASNL~\cite{PointASNL2020} & -    & 86.1 \\
        AdaptConv~\cite{Adaptive2021}  & 83.4 & 86.4 \\
        PT~\cite{point-trans2021}          & 83.7 & 86.6 \\    
        PointMLP~\cite{PointMLP2022}   & 84.6 & 86.1 \\
        KPConv~\cite{KPConv2019}       & 85.1 & 86.4 \\
        \midrule[0.3pt]
        PointViG (Ours)                 & 83.2 & 85.9 \\        
        \bottomrule[1pt]
    \end{tabular}
    \vspace{5pt}
    \caption{Part segmentation results for the ShapeNetPart dataset.}
    \label{tab-partseg}
\end{table}

\subsection{Semantic Segmentation}  \label{experiment-semantic-seg}
The semantic segmentation experiment is conducted on the S3DIS dataset, a large-scale indoor scene dataset encompassing point clouds from six areas, totaling 271 rooms. Each point is characterized by $xyz$ coordinates and RGB features, annotated with a semantic label from 13 categories. The challenging Area-5 is selected for testing, while the remaining areas serve as training data. Our data preprocessing employs the entire room as input.

\textbf{The relevant settings of adaptive dilated graph convolution.} In S3DIS segmentation experiments, we have introduced adaptive dilated graph convolutions based on the PointViG Module. Given the highly uneven distribution of point density in large-scale scenes, to ensure a consistent perceptual field for each subgraph, we employed Ball-Query, denoted by $r$ for the radius of the sphere and $k$ for the number of sparse neighbor nodes sampled within the subgraph. During actual operation, to achieve tensor alignment, the Ball-Query algorithm requires the additional specification of a parameter $m$. If the actual number of points within the sphere is less than $m$, the algorithm ensures that the number of returned points equals $m$ through replication and padding operations. In the adaptive sparse sampling phase, to prevent these invalid filling points in the subgraph from being selected, we have modified the Ball-Query algorithm by introducing a masking operation. This enhancement ensures that the influence of these filling points is shielded during the process of adaptively sampling neighbor nodes. We set $r$ to 0.2, $m$ to 64 and $k$ to 32.

\textbf{The pertinent configurations of the network architecture.} The encoder comprises five stages, with the downsampling ratio and the number of blocks for each stage specified as $(1,4,4,4,4)$ and $(1,2,3,2,2)$, respectively, as depicted in Fig. \ref{Seg_network}. As the neural network progresses in depth, there is a gradual transition from low-level geometric features to more intricate high-level semantic features. To address shallow stages, we employ the grouping-MLP-pooling inference schedule (Stage 1) for efficiency. In deeper stages (Stage 2 to 5), we introduce the proposed ADGC as fundamental building blocks to aggregate local features within the feature space. 

\textbf{Results.} Evaluation metrics, namely mIoU, mAcc, and OA, are chosen for performance assessment. The experimental results presented in Tab. \ref{tab-s3dis} demonstrate that, among the methods listed, PointViG outperforms various graph-based approaches by a substantial margin. In a comprehensive evaluation, considering both the number of parameters and performance, PointViG achieves the optimal trade-off between performance and model complexity.

\begin{table}[ht]
    \centering
    \footnotesize
    %\scriptsize
    \setlength{\tabcolsep}{2.0mm}
    \begin{tabular}{l|c|ccc}
        \toprule[1pt]
        \textbf{Method}                      & Params   & mIoU  & OA    & mAcc  \\
        \midrule[0.3pt]
        PointNet~\cite{PointNet2017}         & 3.6 M    & 41.1  & –     & 49.0  \\
        SegCloud~\cite{SEGCloud2017}         &  -       & 48.9  & –     & 57.4  \\
        PointCNN~\cite{PointCNN2018}         & \textbf{0.6 M}& 57.3& 85.9 & 63.9  \\
        SPG~\cite{SPG2018}                   & -        & 58.0  & 86.4  & -     \\
        PCCN~\cite{PCNN2018}                 & -        & 58.3  & –     & 67.0  \\
        PointWeb~\cite{PointWeb2019}         & -        & 60.3  & 87.0  & 66.6  \\
        HPEIN~\cite{HPEIN2019}               & -        & 61.9  & 87.2  & 68.3  \\
        PointASNL~\cite{PointASNL2020}       & -        & 62.6  & 87.7  & 68.5  \\
        GACNet~\cite{Graph2019}              & -        & 62.8  & 87.7  & -     \\
        KPConv~\cite{KPConv2019}             & 15.0 M   & 67.1  & –     & 72.8 \\
        PointNext-B~\cite{PointNext2022}     & 3.8 M    & 67.3  & 89.4  & 73.7  \\
        AdaptConv~\cite{Adaptive2021}        & -        & 67.9  & 90.0  & 73.2  \\ 
        PointNext-L~\cite{PointNext2022}     & 7.1 M    & 69.0  & 90.0  & 75.3  \\
        PT~\cite{point-trans2021}                & -        & 70.4  & 90.8  & 76.5  \\
        PointNext-XL~\cite{PointNext2022}    & 41.6 M   & 70.5  & 90.6  & 76.8  \\
        \midrule[0.3pt]
        PointViG (Ours)                       & 5.3 M    & \textbf{71.7}& \textbf{90.8} & \textbf{78.9} \\ 
        \bottomrule[1pt]
    \end{tabular}
    \vspace{5pt}
    \caption{Semantic segmentation results for the S3DIS dataset, evaluated on Area-5 (\%).}
    \label{tab-s3dis}
\end{table}

\subsection{Complexity Analysis}
Tab. \ref{tab-complexity} details the space complexity (number of parameters) and time complexity (floating-point operations per sample) for typical models in the context of the ModelNet40 classification task. Specifically, DGCNN~\cite{PointCNN2018}, DeepGCN~\cite{DeepGCNs2021}, GAPointNet~\cite{GAPointNet2021}, and PointWavelet~\cite{PointWavelet2023} are classified as graph-based approaches within this table. PointViG demonstrates superior accuracy compared to all other methods. Regarding model size, PointViG has the fewest parameters, second only to PointCNN~\cite{PointCNN2018}. Moreover, PointViG also exhibits significantly lower FLOPs than other models, affirming its computational efficiency, as depicted in Fig. \ref{complexity}.
        
\begin{table}[ht]
    \centering
    \newcommand{\tabincell}[2]{\begin{tabular}{@{}#1@{}}#2\end{tabular}}
    \footnotesize
    \setlength{\tabcolsep}{2.0mm}
    \begin{tabular}{l|cccc}
        \toprule[1pt]
        %                   & Params. & FLOPs  & mAcc   & OA       \\
        % \textbf{Method} 	& (M)     & (G)    & (\%)   & (\%)     \\  
        \textbf{Method}     & \tabincell{c}{Params.\\(M)}   & \tabincell{c}{FLOPs\\(G)}
                            & \tabincell{c}{mAcc\\(\%)}     & \tabincell{c}{OA\\(\%)} \\  
        \midrule[0.3pt]
        % \midrule[0.3pt]
        PointNet~\cite{PointNet2017}					& 3.5           & 0.9   & 86.2  & 89.2 \\
        PointNet++(MSG)~\cite{PointNetplus2017}			& 1.7           & 4.1   & -     & 91.9 \\
        PointCNN~\cite{PointCNN2018}					& \textbf{0.6}  & -     & 88.1  & 92.2 \\
        DGCNN~\cite{DGCNN2019}					        & 1.8           & 4.8   & 90.2  & 92.9 \\
        GAPointNet~\cite{GAPointNet2021}                & 22.9          & -     & 89.7  & 92.4 \\
        DeepGCN~\cite{DeepGCNs2021}                     & 2.2           & 3.9   & 90.9  & 93.6 \\
        PointNext-S~\cite{PointNext2022}                & 4.5           & 6.5   & 90.9  & 93.7 \\
        PointMLP~\cite{PointMLP2022}                    & 13.2          & 31.3  & 90.9  & 94.1 \\
        PointWavelet-L~\cite{PointWavelet2023}          & 58.4          & 39.2  & 91.1  & \textbf{94.3}\\
        \midrule[0.3pt]
        PointViG (Ours)                                  & 1.5           & \textbf{0.6}   & \textbf{91.2}  & \textbf{94.3} \\
        \bottomrule[1pt]
    \end{tabular}
    \vspace{5pt}
    \caption{The complexity of ModelNet40 classification. The optimal scores in each column highlighted in bold (M: $10^{6}$, G: $10^{9}$).}
    \label{tab-complexity}
\end{table}

\subsection{Feature Diversity and Over-Smoothing} \label{diversity}
In this section, we assess the diversity of output features at each layer within the PointViG Module (refer to Fig. \ref{PV-module}). The metric for feature diversity is expressed as follows:

\begin{equation}
    \label{eq-diversity}
    \text{Diver}(X)=\frac{\left \| X-\overline{X} \right \| _{2}}{M\times N\times d} 
\end{equation}

In Eq. \ref{eq-diversity}, where $X$ and $\overline{X}$ represent the output features of all samples in the test set and their means at a specific layer, and $M$, $N$, and $d$ denote the number of samples, points, and feature dimensionality, respectively. Consequently, $M\times N\times d$ represents the total number of elements in $X$. Essentially, when features in $X$ converge, the presence of high redundancy results in a low $\text{Diver}(X)$. 

\begin{figure}[ht]
    \centering
    \includegraphics[width=0.5\linewidth]{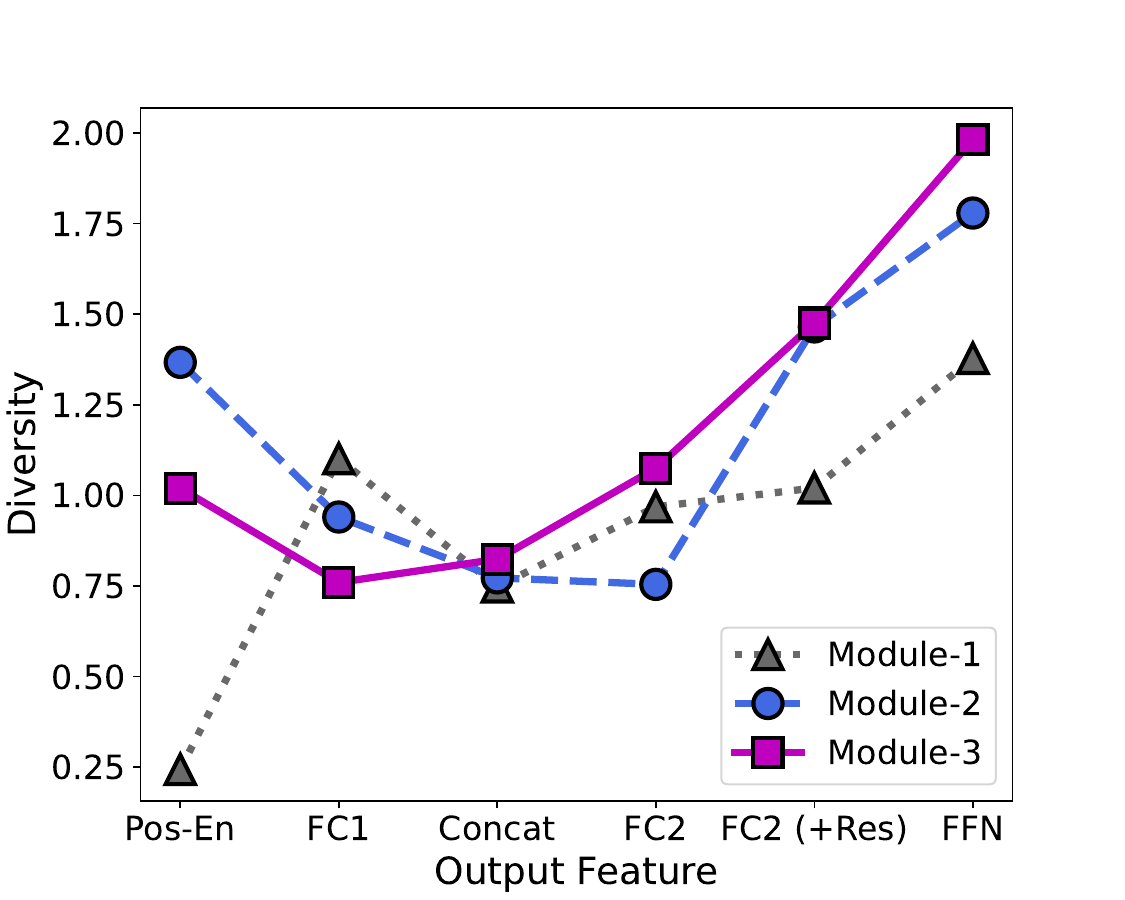}
    \caption{Illustration of the feature diversity output by each layer within the three PointViG Modules utilized in classification models.}
    \label{graph-feature-diversity}
\end{figure}

The classification model in this study consists of three stages, each corresponding to a PointViG Module, and Fig. \ref{graph-feature-diversity} depicts three curves corresponding to Module-0, Module-1, and Module-2, respectively. Horizontal axis labels in the graph indicate layers within the PointViG Module. In these annotations, ``FC2 (+Res)'' signifies the outcome after integrating the output of FC$_2$ and the feature passed by the skip connection. The portion of the PointViG Module situated between FC$_1$ and FC$_2$, constituting the basic graph convolution operation, corresponds to the GraphConv Kernel in PointViG Module. Consequently, the variation in feature diversity from FC$_1$ to the ``Concat'' layer depicted in Fig. \ref{graph-feature-diversity} effectively mirrors the influence of basic graph convolution on feature diversity. 

We conducted a comprehensive analysis of Fig. \ref{graph-feature-diversity} from multiple perspectives, detailed below:
\begin{itemize}
\item The output of the GraphConv Kernel, specifically the "Concat" layer, shows the lowest feature diversity across the curves, except for the "Pos-En" layer in Module-0. This is attributed to over-smoothing in graph convolution, caused by aggregating features with high semantic relevance in a high-dimensional space, leading to the convergence of node features.

\item For FC$_1$, Modules 2 and 3 exhibit noteworthy input feature diversity due to their close association with preceding module outputs. However, the over-smoothing effect and back-propagation reduce feature diversity for Modules 2 and 3 through the FC$_1$ layer. In contrast, Module-0, distinguished from Modules 2 and 3, incorporates low-level geometric features with minimal feature diversity post Position Encoding in Module-1. The FC$_1$ in Module-1 significantly increases feature diversity, highlighting FC$_1$'s pivotal role in the PointViG Module.

\item After the "Concat" layer, a discernible improvement in feature diversity is observed, affirming the substantial contributions of FC$_2$, skip connections, and FFN in enhancing feature diversity.

\item A comparative analysis of the output from each FFN block indicates a gradual increase in feature diversity for each module with increasing network depth.
\end{itemize}

Synthesizing these findings with prior analyses underscores the crucial role of the components in the PointViG Module in alleviating over-smoothing in graph convolution. This establishes the PointViG Module as an effective graph convolution operator.

\subsection{The Effect of Adaptive Dilated Graph Convolution}
The proposed adaptive dilated graph convolution adaptively searches for neighboring nodes in dilated perspectives based on semantic associations. In contrast, conventional dilated convolution typically specifies the neighborhood uniformly or randomly. In the semantic segmentation model, we set $r=0.2$, $m=64$, and $k=32$. Tab. \ref{tab-dilation} presents the performance comparison of the semantic segmentation task when using dilated graph convolution in uniform, random, and adaptive manners. The segmentation model achieves the best performance with adaptive dilated graph convolution.

\begin{table}[ht]
    \centering
    \newcommand{\tabincell}[2]{\begin{tabular}{@{}#1@{}}#2\end{tabular}}
    \small
    %\footnotesize
    \setlength{\tabcolsep}{7.2mm}
    \begin{tabular}{l|ccc}
        \toprule[1pt]
        Dilation &  \tabincell{c}{Uniform} &  \tabincell{c}{Random} & \tabincell{c}{Adaptive} \\
        \midrule[0.3pt]
        % \midrule[0.3pt]
        mIoU (\%)		    & 69.0    & 67.4    & 71.7 \\
        \bottomrule[1pt]

    \end{tabular}
    \vspace{5pt}
    \caption{Comparative performance analysis of dilated graph convolution with different sparse sampling strategies for S3DIS (Area-5) segmentation.}
    \label{tab-dilation}
\end{table}

\subsection{Ablation Studies}

\textbf{The design of PointViG Module.} Ablation experiments are conducted to assess the impact of these components in the PointViG Module on classification tasks. The experimental results are presented in Tab. \ref{tab-vig}. Abbreviations such as "Pos-En," "FC${_1}$," "FC${_2}$," "FFN," and "concat" represent position encoding, the FC${_1}$ layer, the FC${_2}$ layer, the FFN block, and the concatenation operation between the central node and the local feature in Fig. \ref{PV-module}. Models 1, 2 and 3 involve the removal of position encoding, FC layers, and FFN, respectively, showcasing varying degrees of performance degradation. In Model-4, FC layers and FFN are removed, while in Model-5, the PointViG Module is reduced to a pure GraphConv Kernel. The results of Models 4 and 5 underscore that direct utilization of the GraphConv Kernel for point cloud classification leads to substantial performance degradation. Model-6 explores the concatenation operation in the GraphConv Kernel, confirming its ability to supplement the central node's features to the graph convolution output and strengthen feature characterization. The performance discrepancies between these ablation models and the PointViG model echo the analysis of feature diversity presented in Sec. \ref{diversity}.

\begin{table}[ht]
    \centering
    \small
    %\footnotesize
    \setlength{\tabcolsep}{3.3mm}
    \begin{tabular}{l|cccccc}
        \toprule[1pt]
        % \toprule
        Model    &Pos-En      &FC1         &FC2         &FFN          &Concat      & OA(\%) \\
        \midrule[0.3pt]
        % \midrule[0.3pt]
        Model-1  &\ding{53}   &\ding{52}   &\ding{52}   &\ding{52}    &\ding{52}   & 92.1   \\
        Model-2  &\ding{52}   &\ding{53}   &\ding{53}   &\ding{52}    &\ding{52}   & 93.1   \\
        Model-3  &\ding{52}   &\ding{52}   &\ding{52}   &\ding{53}    &\ding{52}   & 93.3   \\
        Model-4  &\ding{52}   &\ding{53}   &\ding{53}   &\ding{53}    &\ding{52}   & 92.6   \\
        Model-5  &\ding{53}   &\ding{53}   &\ding{53}   &\ding{53}    &\ding{52}   & 90.8   \\
        Model-6  &\ding{52}   &\ding{52}   &\ding{52}   &\ding{52}    &\ding{53}   & 93.5   \\
        \midrule[0.3pt]
        PointViG (Ours)    
                 &\ding{52}   &\ding{52}   &\ding{52}   &\ding{52}    &\ding{52}   & 94.3   \\         
        \bottomrule[1pt]
        % \bottomrule
    \end{tabular}
    \vspace{5pt}
    \caption{Ablation study of the PointViG Module design for classification.}
    \label{tab-vig}
\end{table}

\textbf{Comparative analysis of PointViG and other graph convolution operators.} We evaluate prominent variations of graph convolution kernels in Tab. \ref{tab-operator}. Instead of directly employing these kernels as the fundamental module, we replace the GraphConv Kernel in the proposed PointViG Module with these kernels for a fair comparison. The PointViG Module demonstrates superior performance compared to the other kernels, suggesting its superior ability to aggregate point cloud features. 

\begin{table}[ht]
    \centering
    \small
    %\footnotesize
    \setlength{\tabcolsep}{6.6mm}
    \begin{tabular}{l|cc}
        \toprule[1pt]
        kernel                 & mAcc(\%)  & OA(\%) \\
        \midrule[0.3pt]
        % \midrule[0.3pt]
        GIN~\cite{GIN2019}          & 88.9  & 92.5 \\
        MRGCN~\cite{DeepGCNs2021}   & 90.1  & 93.1 \\
        EdgeConv~\cite{DGCNN2019}   & 90.6  & 93.6 \\
        GraphSAGE~\cite{SAGE2017}   & 89.5  & 93.2 \\
        \midrule[0.3pt]
        PointViG (Ours)             & 91.2	& 94.3 \\
        \bottomrule[1pt]
    \end{tabular}
    \vspace{5pt}
    \caption{Embedding of various graph convolutional kernels within the PointViG Module for a fair performance comparison within the same framework on ModelNet40 classification.}
    \label{tab-operator}
\end{table}

\textbf{The neighborhood size for ModelNet40 classification.} In the design of the classification model, we simplify by setting the number of neighborhood nodes ($k$) to a consistent value at each stage. We assess the influence of different $k$ on performance. The experimental results shown in Tab. \ref{tab-k} indicate that the classification task reaches its peak performance when the parameter $k$ is set to 16.

\begin{table}[ht]
    \centering
    \small
    %\footnotesize
    \setlength{\tabcolsep}{4.5mm}
    \begin{tabular}{c|cccc}
        \toprule[1pt]
        $k$                     & 4  	& 8  	& 12  	& 16 \\
        \midrule[0.3pt]
        % \midrule[0.3pt]
       % mAcc(\%)				& 89.8	& 91.2  & 91.6  & 90.3 \\
         OA(\%) 					& 93.1 	& 94.3  & 93.6  & 93.4 \\ 
        \bottomrule[1pt]
    \end{tabular}
    \vspace{5pt}
    \caption{ModelNet40 classification results with varying numbers ($k$) of nearest neighbors.}
    \label{tab-k}
\end{table}

\textbf{The effect of neighhood size ($k$) on S3DIS segmentation.} 
Tab. \ref{tab-segmentation-ADGC} compares the performance of S3DIS semantic segmentation with different $k$ values when $r$ is fixed at 0.2. The model achieves optimal performance when $k=32$. This observation indicates that treating all nodes in the subgraph as neighborhood nodes does not improve performance. The adaptive dilated graph convolution strategy demonstrates its effectiveness by adaptively selecting nodes with the strongest semantic associations.

\begin{table}[ht]
    \centering
    \small
    %\footnotesize
    \setlength{\tabcolsep}{2.2mm}
    \begin{tabular}{c|cccccc}
        \toprule[1pt]
      $k$                 & 8    & 16    & 24    & 32    & 40    & 48  \\
        \midrule[0.3pt]
        % \midrule[0.3pt]
        mIoU (\%)		    & 69.0    & 70.5    & 71.4   & 71.7  & 70.4  & 70.3\\
        \bottomrule[1pt]

    \end{tabular}
    \vspace{5pt}
    \caption{Segmentation results for S3DIS (Area-5) with varying numbers of neighboring nodes.}
    \label{tab-segmentation-ADGC}
\end{table}

\textbf{The effect of radius ($r$) on S3DIS segmentation.} 
In S3DIS segmentation, we employ the Ball-Query, where the parameters $r$ and $k$ define the subgraph range and the number of neighboring nodes. Increasing $r$ expands the perceptual field of the graph convolution without a corresponding increase in computational cost. We maintain a constant number of neighboring nodes at 32 and explore the impact of varied $r$ values on segmentation performance. The results, depicted in Tab. \ref{tab-segmentation-r}, indicate optimal performance when $r$ is set to 0.2. 

In Tab. \ref{tab-segmentation-r}, when $r = 0.1$, the subgraph is small, fostering strong semantic associations among nodes within the subgraph. Typically, $r$ is set to 0.1 in PointNext~\cite{PointNext2022}. As $r$ increases to 0.12 and 0.14, the subgraph range expands, but perceptual field expansion remains limited, resulting in weakened semantic associations between valid points compared to $r = 0.1$. As $r$ continues to increase to larger values, the perceptual field undergoes a dramatic expansion. The central node gains enhanced flexibility in searching for neighboring nodes within a broader range, establishing semantic associations with distant nodes, thereby contributing to further performance improvement.

\begin{table}[ht]
    \centering
    \small
    %\footnotesize
    \setlength{\tabcolsep}{2.2mm}
    \begin{tabular}{c|cccccc}
        \toprule[1pt]
          $r$                 & 0.10    & 0.12    & 0.14   & 0.16  & 0.18  & 0.20  \\
        \midrule[0.3pt]
        % \midrule[0.3pt]
        mIoU (\%)		    & 70.0    & 69.1    & 69.6   & 70.7  & 70.9  & 71.7   \\
        \bottomrule[1pt]

    \end{tabular}
    \vspace{5pt}
    \caption{Semantic segmentation results on S3DIS Dataset (Area-5) with varying radii of Ball-Query.}
    \label{tab-segmentation-r}
\end{table}

\begin{figure}[ht]
    \centering
    \includegraphics[width=0.5\linewidth]{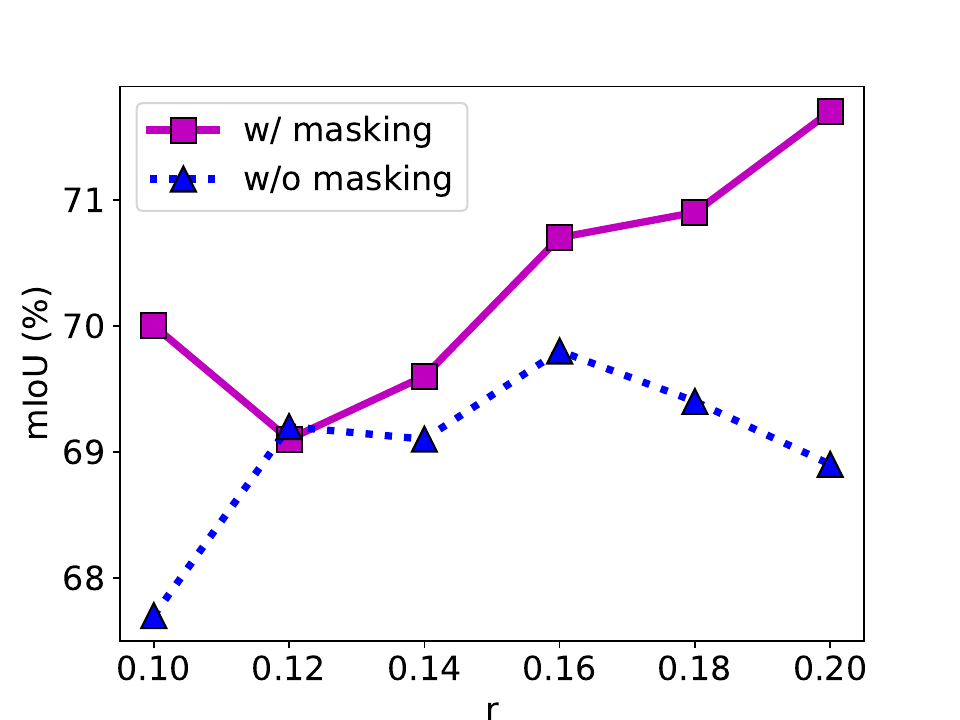}
    \caption{Segmentation results of S3DIS (Area-5) with and without the masking operation in adaptive dilated graph convolution with varied radii of Ball-Query.}
    \label{graph-masking}
\end{figure}

\textbf{The effect of masking operation on S3DIS segmentation.}\label{mask-ablation}
The S3DIS segmentation model employs Ball-Query, leading to the inclusion of invalid points in the acquired subgraphs obtained by filling operation. The presence of invalid points affects the adaptive search for neighborhood nodes. To mitigate this issue, we enhance the Ball-Query algorithm to output not only the subgraph but also the corresponding mask. During adaptive dilated graph convolution, we utilize the mask flag to mitigate the impact of duplicate points when searching for neighborhood nodes in the subgraph. To validate the effectiveness of the masking operation, we compare the performance of the PointViG framework both with and without the masking operation. The masking operation overall yields a significant performance gain. The complex distribution of points leads to significant variations in both the perceived fields and interactions among adjacent nodes when the parameter $r$ undergoes alterations. Consequently, accurately predicting the trend of the curves depicted in Fig. \ref{graph-masking} becomes challenging. Upon an overall comparison of the two curves, it can be observed that, in general, the performance of adaptive dilated graph convolution is superior when employing a masking operation compared to when it is not used. Moreover, the performance gap becomes more pronounced at larger values of radius $r$.

\subsection{Visualization}
\subsubsection{Visualization of t-SNE Feature Clustering} \label{sec-vis-tsne}
\begin{figure}[htbp]
    \centering
    \includegraphics[width=\linewidth]{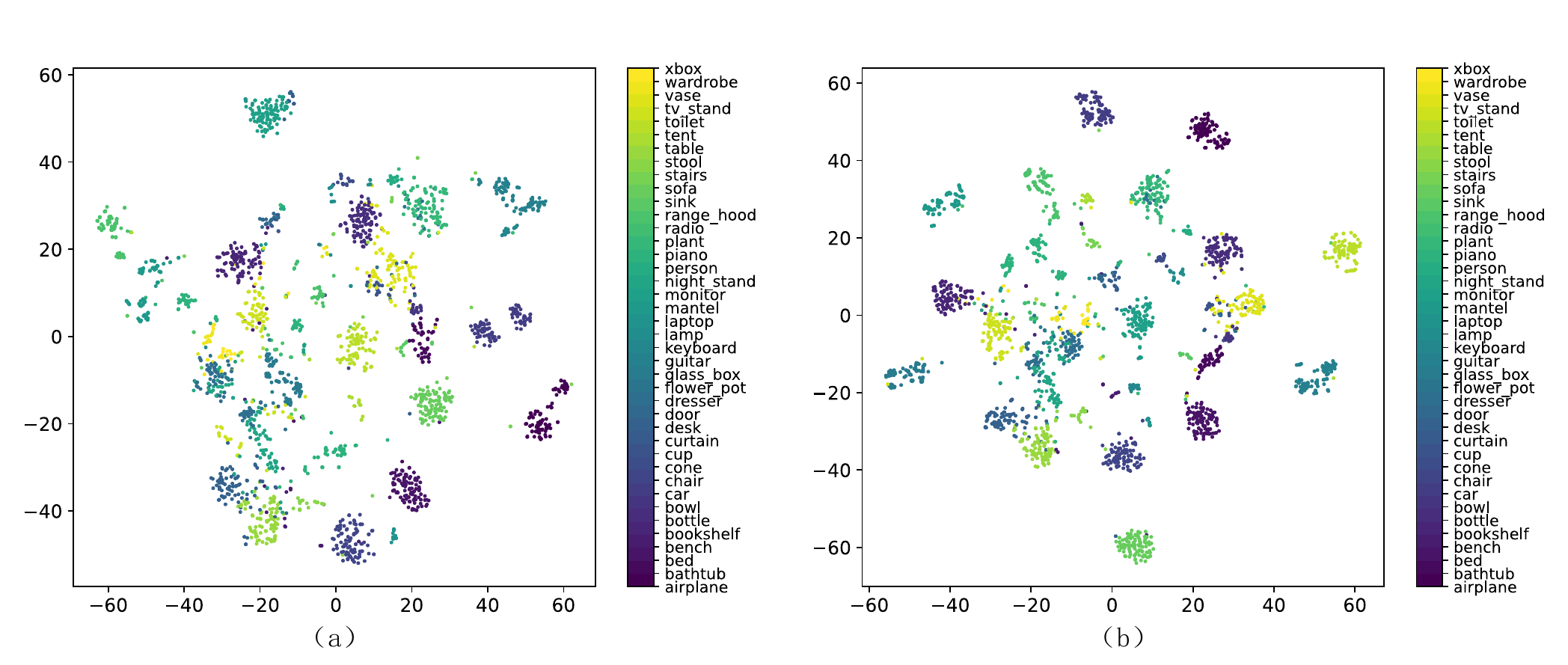}
    \caption{t-SNE visualization of encoded features for (a) DGCNN~\cite{DGCNN2019} and (b) PointViG on the ModelNet40 test set.}
    \label{figure-tsne-feature}
\end{figure}

To further evaluate the feature representation capabilities of the proposed PointViG model, we employed t-SNE (t-Distributed Stochastic Neighbor Embedding) for the downscaling and clustering analysis of encoder features. This technique was applied to the ModelNet40 test set, enabling a comparative visual analysis between PointViG and the widely-used DGCNN model~\cite{DGCNN2019}, which also adopts a GNN-based architecture. The clustering results are presented in Fig. \ref{figure-tsne-feature}(a) for DGCNN~\cite{DGCNN2019} and Fig. \ref{figure-tsne-feature}(b) for PointViG.

The t-SNE visualizations of DGCNN's features reveal a pattern where the majority of clusters are not clearly separated, indicating that DGCNN~\cite{DGCNN2019} struggles to maintain distinct and independent feature representations. This suggests a limited capacity in the DGCNN~\cite{DGCNN2019} model to differentiate between similar point cloud structures within the dataset. The lack of well-defined boundaries highlights potential challenges in achieving robust classification performance.

In contrast, the feature space of PointViG, as visualized through t-SNE, demonstrates a substantial improvement in cluster separability. The clusters are independently distributed with minimal overlap, and most are tightly grouped, with clear boundaries between them. This pattern reflects PointViG’s enhanced ability to capture diverse geometric features from the point cloud data, leading to a more discriminative and robust feature representation. Such results suggest that the PointViG model is capable of better generalization and classification performance compared to DGCNN~\cite{DGCNN2019}.

\subsubsection{Visualization of Neighboring Nodes} \label{sec-vis-neigh}

\begin{figure*}[bp]
    \centering
    \includegraphics[width=0.9\linewidth]{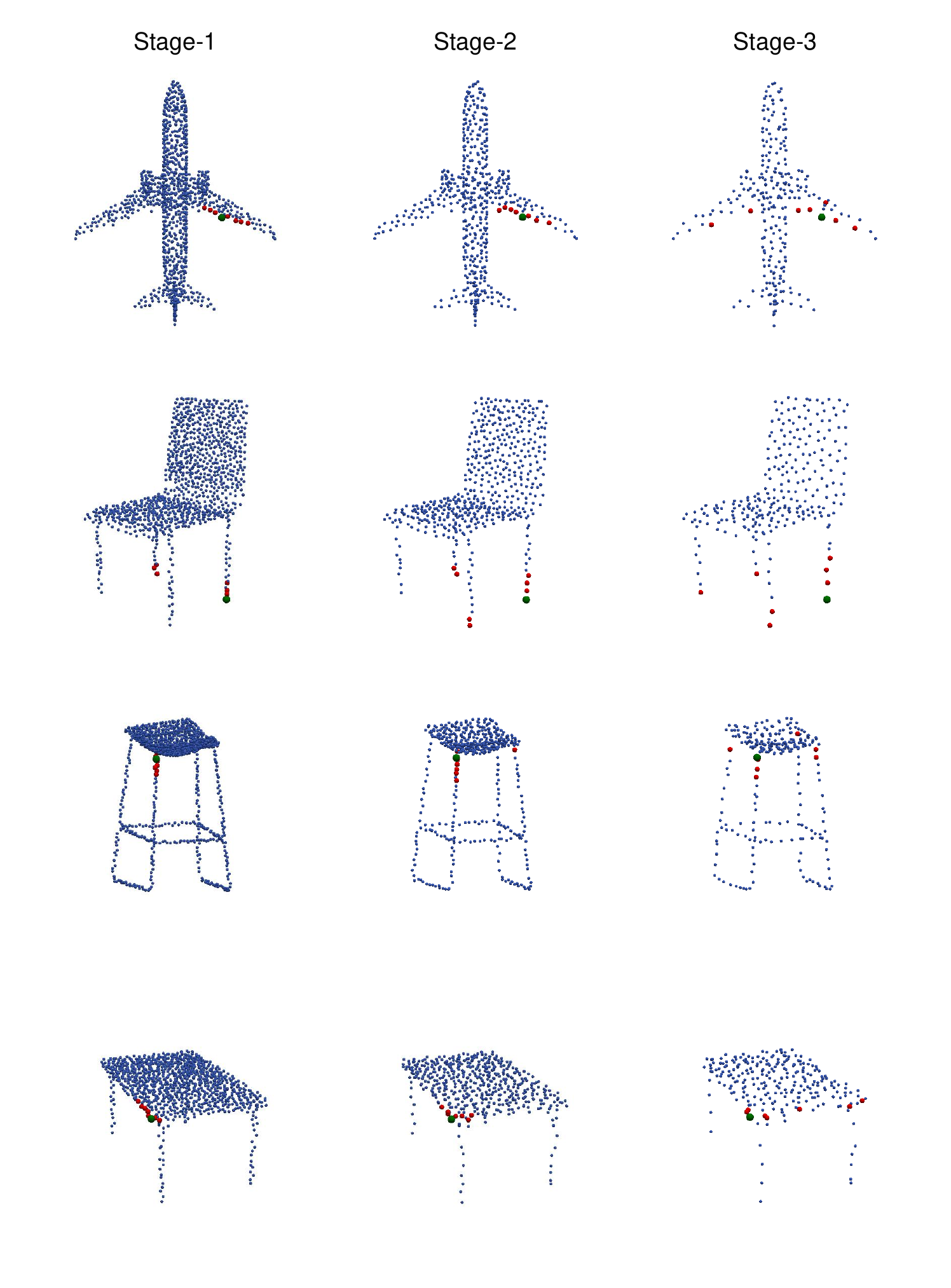}
    \caption{\textbf{Visualization of graph node neighborhoods}. Each row represents a sample from the ModelNet40 test set. The three columns correspond to the three stages of the classification model encoder. Green dots represent the central node, while its neighboring nodes are denoted by red dots.}
    \label{graph-vis-neigh-node}
\end{figure*}

In Fig. \ref{graph-vis-neigh-node}, we depict the graph node neighborhoods for samples in the ModelNet40 test set. The three columns in the illustration correspond to the stages (Stage-1 to Stage-3) of the classification model encoder, representing the network's progressive deepening. The downsampling operation in each stage leads to a gradual sparsification of the point cloud. Initially constrained to local regions in the shallow stage, neighbor nodes progressively transcend spatial constraints, exploring globally for nodes with high semantic relevance.

Illustratively, in the first row depicting an airplane, the neighbor nodes extend from the local area of the right wing edges to encompass the entire wingspan. In the second row featuring a chair, a specific point on the legs expands from an initial distribution involving two legs to three and four, capturing the crucial structural information of the chair having four symmetric legs. The third-row stool exhibits a similar pattern. The fourth-row table illustrates a central point at one corner gradually expanding its neighboring nodes from one corner to the other along the table edge.

These examples affirm the effectiveness of the PointViG Module proposed in this paper in capturing neighboring nodes with semantic associations, facilitating feature aggregation for enhanced target identification.

\subsubsection{Visualization of Part Segmentation} \label{sec-vis-part-seg}

\begin{figure*}[bp]
    \centering
    \includegraphics[width=0.9\linewidth]{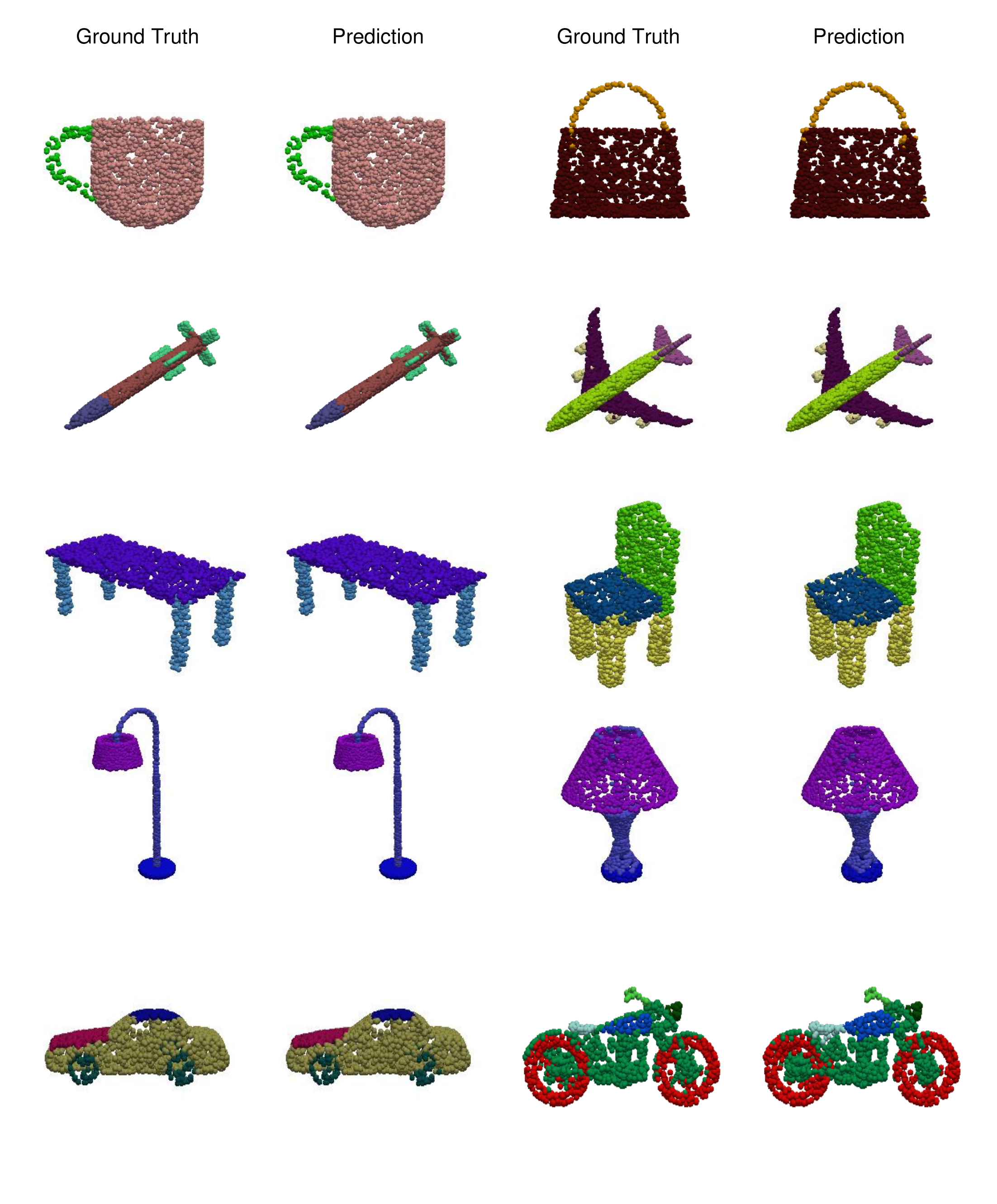}
    \caption{\textbf{Visualization of part segmentation results}. For each sample, the left side displays the ground truth, while the right side illustrates the corresponding prediction.}
    \label{graph-vis-part}
\end{figure*}

A visualization of the part segmentation experiment results is presented in Fig. \ref{graph-vis-part}. Generally, the segmentation performance is superior for targets with a simple structure. However, for targets characterized by a complex structure, deviations in predictions are observed at points where different parts are combined. This is evident in specific instances such as the junction of the tail and body of the rocket in the second row, the fusion of the rear wheel and car body, and the combination of the fuel tank and Motorbike body in the fourth row.

\subsubsection{Visualization of Semantic Segmentation} \label{sec-vis-semantic-seg}

Fig. \ref{graph-vis-s3dis} displays the results of the semantic segmentation for S3DIS (Area-5). In general, the PointViG model exhibits a relatively accurate performance in the segmentation task. The segmentation inaccuracies observed in Fig. \ref{graph-vis-s3dis} can be primarily attributed to the following factors:

\begin{figure*}[bp]
    \centering
    \includegraphics[width=0.95\linewidth]{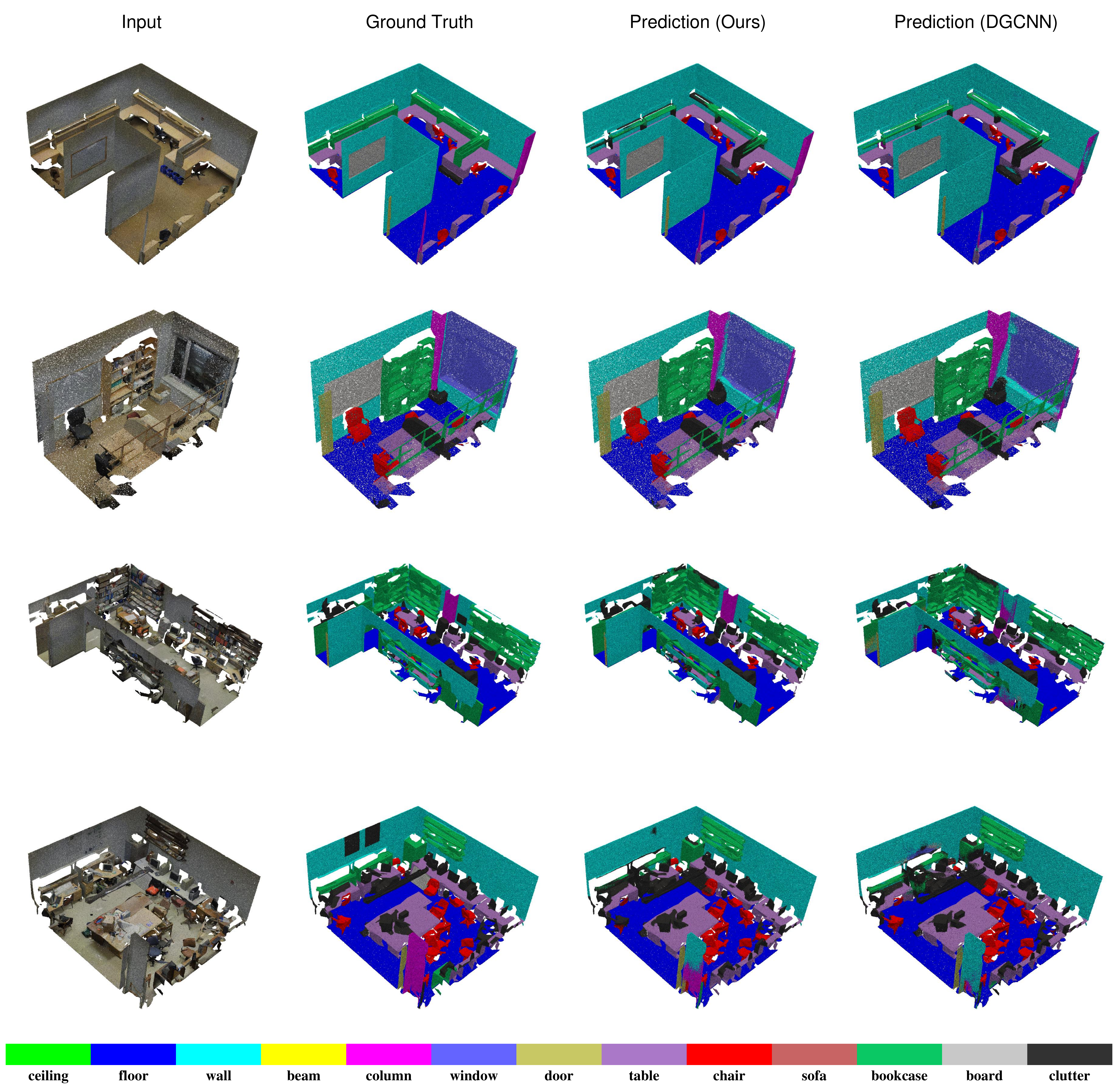}
    \caption{\textbf{Visualization of semantic segmentation results in S3DIS (Area-5)}. The four columns depict, from left to right, the input data, ground truth, prediction (Ours) and prediction (DGCNN~\cite{DGCNN2019}). To enhance the visualization, we excluded the ceiling and a portion of the walls.}
    \label{graph-vis-s3dis}
\end{figure*}

\begin{itemize}
\item Small-sized targets in one or more dimensions.
The elongated bookcase in the first-row sample scene poses a challenge in capturing features accurately and determining semantic boundaries, leading to the incorrect identification of some points as clutter. In contrast, the larger bookcase in the second-row sample scene yields more accurate identification.

\item Semantically related and neighboring targets.
In the sample scene of the fourth row, the model incorrectly labels the upper part of a column as a wall, influenced by the strong semantic correlation and frequent co-occurrence between the column and the wall. The visual similarity between the two complicates accurate differentiation for the model.

\item Targets with obscure geometric and color features.
The small rectangular clutter area between two bookcases in the third-row sample scene and the adjacent rectangular clutter areas on the wall in the fourth-row sample scene are misclassified as walls. These clutter regions closely resemble the background walls in both color and geometric features, presenting a challenge for the model in differentiation.

\end{itemize}

\section{Conclusion}
In this paper, we introduce a novel computationally efficient graph convolutional operator, PointViG, complemented by an adaptive dilated graph convolution strategy tailored for large-scale point cloud scenarios. Leveraging these technologies, we construct an effective framework for point cloud analysis. This framework achieves performance comparable to SOTA models in tasks such as classification and segmentation while significantly reducing complexity, thereby achieving an optimal balance between performance and complexity. PointViG framework provides a viable solution for deploying point cloud analysis models in resource-constrained environments. Although PointViG focuses primarily on point cloud analysis, its potential for application in other fields involving graph data analysis remains largely untapped. In future research, we plan to incorporate model compression techniques to further reduce model complexity and explore the possibility of extending PointViG to other domains.

\bibliographystyle{elsarticle-num}
\bibliography{reference}% common bib file

\begin{thebibliography}{10}
\expandafter\ifx\csname url\endcsname\relax
  \def\url#1{\texttt{#1}}\fi
\expandafter\ifx\csname urlprefix\endcsname\relax\def\urlprefix{URL }\fi
\expandafter\ifx\csname href\endcsname\relax
  \def\href#1#2{#2} \def\path#1{#1}\fi

\bibitem{VoxNet2015}
D.~Maturana, S.~Scherer, Voxnet: A 3d convolutional neural network for real-time object recognition, in: 2015 IEEE/RSJ International Conference on Intelligent Robots and Systems (IROS), 2015, pp. 922--928.
\newblock \href {https://doi.org/10.1109/IROS.2015.7353481} {\path{doi:10.1109/IROS.2015.7353481}}.

\bibitem{SPLATNet2018}
H.~Su, V.~Jampani, D.~Sun, S.~Maji, E.~Kalogerakis, M.-H. Yang, J.~Kautz, Splatnet: Sparse lattice networks for point cloud processing, in: 2018 IEEE/CVF Conference on Computer Vision and Pattern Recognition, 2018, pp. 2530--2539.
\newblock \href {https://doi.org/10.1109/CVPR.2018.00268} {\path{doi:10.1109/CVPR.2018.00268}}.

\bibitem{OctNet2017}
G.~Riegler, A.~O. Ulusoy, A.~Geiger, Octnet: Learning deep 3d representations at high resolutions, in: 2017 IEEE Conference on Computer Vision and Pattern Recognition (CVPR), 2017, pp. 6620--6629.
\newblock \href {https://doi.org/10.1109/CVPR.2017.701} {\path{doi:10.1109/CVPR.2017.701}}.

\bibitem{2015Multi}
H.~Su, S.~Maji, E.~Kalogerakis, E.~Learned-Miller, Multi-view convolutional neural networks for 3d shape recognition, in: 2015 IEEE International Conference on Computer Vision (ICCV), 2015, pp. 945--953.
\newblock \href {https://doi.org/10.1109/ICCV.2015.114} {\path{doi:10.1109/ICCV.2015.114}}.

\bibitem{multi2021}
W.~Wang, T.~Wang, Y.~Cai, Multi-view attention-convolution pooling network for 3d point cloud classification, Applied Intelligence (2021) 1--12\href {https://doi.org/https://doi.org/10.1007/s10489-021-02840-2} {\path{doi:https://doi.org/10.1007/s10489-021-02840-2}}.

\bibitem{PointNet2017}
R.~Q. Charles, H.~Su, M.~Kaichun, L.~J. Guibas, Pointnet: Deep learning on point sets for 3d classification and segmentation, in: 2017 IEEE Conference on Computer Vision and Pattern Recognition (CVPR), 2017, pp. 77--85.
\newblock \href {https://doi.org/10.1109/CVPR.2017.16} {\path{doi:10.1109/CVPR.2017.16}}.

\bibitem{PointNetplus2017}
C.~R. Qi, L.~Yi, H.~Su, L.~J. Guibas, Pointnet++: Deep hierarchical feature learning on point sets in a metric space, in: Advances in Neural Information Processing Systems, Vol.~30, 2017.

\bibitem{PointConv2019}
W.~Wu, Z.~Qi, L.~Fuxin, Pointconv: Deep convolutional networks on 3d point clouds, in: 2019 IEEE/CVF Conference on Computer Vision and Pattern Recognition (CVPR), 2019, pp. 9613--9622.
\newblock \href {https://doi.org/10.1109/CVPR.2019.00985} {\path{doi:10.1109/CVPR.2019.00985}}.

\bibitem{DGCNN2019}
Y.~Wang, Y.~Sun, Z.~Liu, S.~E. Sarma, M.~M. Bronstein, J.~M. Solomon, Dynamic graph cnn for learning on point clouds, ACM Trans. Graph. 38~(5) (2019).

\bibitem{PT2021}
N.~Engel, V.~Belagiannis, K.~Dietmayer, Point transformer, IEEE Access 9 (2021) 134826--134840.
\newblock \href {https://doi.org/10.1109/ACCESS.2021.3116304} {\path{doi:10.1109/ACCESS.2021.3116304}}.

\bibitem{DeepGCNs2021}
G.~Li, M.~Mueller, G.~Qian, I.~C. Delgadillo~Perez, A.~Abualshour, A.~K. Thabet, B.~Ghanem, Deepgcns: Making gcns go as deep as cnns, IEEE Transactions on Pattern Analysis and Machine Intelligence (2021).

\bibitem{PointWavelet2023}
C.~Wen, J.-L. Long, B.~Yu, D.~Tao, Pointwavelet: Learning in spectral domain for 3d point cloud analysis, ArXiv abs/2302.05201 (2023).

\bibitem{SPG2018}
L.~Landrieu, M.~Simonovsky, Large-scale point cloud semantic segmentation with superpoint graphs, in: 2018 IEEE/CVF Conference on Computer Vision and Pattern Recognition, 2018, pp. 4558--4567.
\newblock \href {https://doi.org/10.1109/CVPR.2018.00479} {\path{doi:10.1109/CVPR.2018.00479}}.

\bibitem{Adaptive2021}
H.~Zhou, Y.~Feng, M.~Fang, M.~Wei, J.~Qin, T.~Lu, Adaptive graph convolution for point cloud analysis, in: 2021 IEEE/CVF International Conference on Computer Vision (ICCV), 2021, pp. 4945--4954.
\newblock \href {https://doi.org/10.1109/ICCV48922.2021.00492} {\path{doi:10.1109/ICCV48922.2021.00492}}.

\bibitem{GNN2005}
M.~Gori, G.~Monfardini, F.~Scarselli, A new model for learning in graph domains, in: Proceedings. 2005 IEEE International Joint Conference on Neural Networks, 2005., Vol.~2, 2005, pp. 729--734 vol. 2.
\newblock \href {https://doi.org/10.1109/IJCNN.2005.1555942} {\path{doi:10.1109/IJCNN.2005.1555942}}.

\bibitem{GNN2009}
F.~Scarselli, M.~Gori, A.~C. Tsoi, M.~Hagenbuchner, G.~Monfardini, The graph neural network model, IEEE Transactions on Neural Networks 20~(1) (2009) 61--80.
\newblock \href {https://doi.org/10.1109/TNN.2008.2005605} {\path{doi:10.1109/TNN.2008.2005605}}.

\bibitem{NN4G2009}
A.~Micheli, Neural network for graphs: A contextual constructive approach, IEEE Transactions on Neural Networks 20~(3) (2009) 498--511.
\newblock \href {https://doi.org/10.1109/TNN.2008.2010350} {\path{doi:10.1109/TNN.2008.2010350}}.

\bibitem{Diffusion2016}
J.~Atwood, D.~Towsley, Diffusion-convolutional neural networks, in: D.~Lee, M.~Sugiyama, U.~Luxburg, I.~Guyon, R.~Garnett (Eds.), Advances in Neural Information Processing Systems, Vol.~29, 2016.

\bibitem{Message2017}
J.~Gilmer, S.~S. Schoenholz, P.~F. Riley, O.~Vinyals, G.~E. Dahl, Neural message passing for quantum chemistry, in: Proceedings of the 34th International Conference on Machine Learning - Volume 70, ICML'17, JMLR.org, 2017, p. 1263–1272.

\bibitem{Learning2016}
M.~Niepert, M.~Ahmed, K.~Kutzkov, Learning convolutional neural networks for graphs, in: Proceedings of the 33rd International Conference on International Conference on Machine Learning - Volume 48, ICML'16, JMLR.org, 2016, p. 2014–2023.

\bibitem{Spectral2013}
J.~Bruna, W.~Zaremba, A.~D. Szlam, Y.~LeCun, Spectral networks and locally connected networks on graphs, CoRR abs/1312.6203 (2013).

\bibitem{Fast2016}
M.~Defferrard, X.~Bresson, P.~Vandergheynst, Convolutional neural networks on graphs with fast localized spectral filtering, in: D.~Lee, M.~Sugiyama, U.~Luxburg, I.~Guyon, R.~Garnett (Eds.), Advances in Neural Information Processing Systems, Vol.~29, 2016.

\bibitem{Clustering2020}
X.~Li, Y.~Hu, Y.~Sun, J.~Hu, J.~Zhang, M.~Qu, A deep graph structured clustering network, IEEE Access 8 (2020) 161727--161738.
\newblock \href {https://doi.org/10.1109/ACCESS.2020.3020192} {\path{doi:10.1109/ACCESS.2020.3020192}}.

\bibitem{Semi2016}
T.~Kipf, M.~Welling, Semi-supervised classification with graph convolutional networks, ArXiv abs/1609.02907 (2016).

\bibitem{Structural2016}
A.~Jain, A.~R. Zamir, S.~Savarese, A.~Saxena, Structural-rnn: Deep learning on spatio-temporal graphs, in: 2016 IEEE Conference on Computer Vision and Pattern Recognition (CVPR), 2016, pp. 5308--5317.
\newblock \href {https://doi.org/10.1109/CVPR.2016.573} {\path{doi:10.1109/CVPR.2016.573}}.

\bibitem{Temporal2018}
S.~Yan, Y.~Xiong, D.~Lin, Spatial temporal graph convolutional networks for skeleton-based action recognition, in: AAAI Conference on Artificial Intelligence, 2018.

\bibitem{Scene2017}
D.~Xu, Y.~Zhu, C.~B. Choy, L.~Fei-Fei, Scene graph generation by iterative message passing, in: 2017 IEEE Conference on Computer Vision and Pattern Recognition (CVPR), 2017, pp. 3097--3106.
\newblock \href {https://doi.org/10.1109/CVPR.2017.330} {\path{doi:10.1109/CVPR.2017.330}}.

\bibitem{Generation2020}
Y.~Guo, J.~Song, L.~Gao, H.~T. Shen, One-shot scene graph generation, in: Proceedings of the 28th ACM International Conference on Multimedia, MM '20, Association for Computing Machinery, New York, NY, USA, 2020, p. 3090–3098.
\newblock \href {https://doi.org/10.1145/3394171.3414025} {\path{doi:10.1145/3394171.3414025}}.

\bibitem{ViG2022}
K.~Han, Y.~Wang, J.~Guo, Y.~Tang, E.~Wu, Vision (gnn): An image is worth graph of nodes, arXiv preprint arXiv:2206.00272 (2022).

\bibitem{PointMLP2022}
X.~Ma, C.~Qin, H.~You, H.~Ran, Y.~Fu, Rethinking network design and local geometry in point cloud: A simple residual mlp framework, in: International Conference on Learning Representations, 2022.

\bibitem{PointNext2022}
G.~Qian, Y.~Li, H.~Peng, J.~Mai, H.~Hammoud, M.~Elhoseiny, B.~Ghanem, Pointnext: Revisiting pointnet++ with improved training and scaling strategies, in: S.~Koyejo, S.~Mohamed, A.~Agarwal, D.~Belgrave, K.~Cho, A.~Oh (Eds.), Advances in Neural Information Processing Systems, Vol.~35, 2022.

\bibitem{2023JSNet++}
L.~Zhao, W.~Tao, Jsnet++: Dynamic filters and pointwise correlation for 3d point cloud instance and semantic segmentation, IEEE Transactions on Circuits and Systems for Video Technology 33~(4) (2023) 1854--1867.
\newblock \href {https://doi.org/10.1109/TCSVT.2022.3218076} {\path{doi:10.1109/TCSVT.2022.3218076}}.

\bibitem{PointCNN2018}
Y.~Li, R.~Bu, M.~Sun, W.~Wu, X.~Di, B.~Chen, Pointcnn: Convolution on x-transformed points, in: S.~Bengio, H.~Wallach, H.~Larochelle, K.~Grauman, N.~Cesa-Bianchi, R.~Garnett (Eds.), Advances in Neural Information Processing Systems, Vol.~31, 2018.

\bibitem{PAConv2021}
M.~Xu, R.~Ding, H.~Zhao, X.~Qi, Paconv: Position adaptive convolution with dynamic kernel assembling on point clouds, in: 2021 IEEE/CVF Conference on Computer Vision and Pattern Recognition (CVPR), 2021, pp. 3172--3181.
\newblock \href {https://doi.org/10.1109/CVPR46437.2021.00319} {\path{doi:10.1109/CVPR46437.2021.00319}}.

\bibitem{SpiderCNN2018}
Y.~Xu, T.~Fan, M.~Xu, L.~Zeng, Y.~Qiao, Spidercnn: Deep learning on point sets with parameterized convolutional filters, in: European Conference on Computer Vision, 2018.

\bibitem{KPConv2019}
H.~Thomas, C.~R. Qi, J.-E. Deschaud, B.~Marcotegui, F.~Goulette, L.~Guibas, Kpconv: Flexible and deformable convolution for point clouds, in: 2019 IEEE/CVF International Conference on Computer Vision (ICCV), 2019, pp. 6410--6419.
\newblock \href {https://doi.org/10.1109/ICCV.2019.00651} {\path{doi:10.1109/ICCV.2019.00651}}.

\bibitem{Set2019}
J.~Lee, Y.~Lee, J.~Kim, A.~Kosiorek, S.~Choi, Y.~W. Teh, Set transformer: A framework for attention-based permutation-invariant neural networks, in: Proceedings of the 36th International Conference on Machine Learning, 2019, pp. 3744--3753.

\bibitem{Point2Sequence2019}
X.~Liu, Z.~Han, Y.-S. Liu, M.~Zwicker, Point2sequence: Learning the shape representation of 3d point clouds with an attention-based sequence to sequence network, in: AAAI Conference on Artificial Intelligence, 2019.

\bibitem{Attentional2018}
S.~Xie, S.~Liu, Z.~Chen, Z.~Tu, Attentional shapecontextnet for point cloud recognition, in: 2018 IEEE/CVF Conference on Computer Vision and Pattern Recognition, 2018, pp. 4606--4615.
\newblock \href {https://doi.org/10.1109/CVPR.2018.00484} {\path{doi:10.1109/CVPR.2018.00484}}.

\bibitem{Modeling2019}
J.~Yang, Q.~Zhang, B.~Ni, L.~Li, J.~Liu, M.~Zhou, Q.~Tian, Modeling point clouds with self-attention and gumbel subset sampling, in: 2019 IEEE/CVF Conference on Computer Vision and Pattern Recognition (CVPR), 2019, pp. 3318--3327.
\newblock \href {https://doi.org/10.1109/CVPR.2019.00344} {\path{doi:10.1109/CVPR.2019.00344}}.

\bibitem{Fast2022}
C.~Park, Y.~Jeong, M.~Cho, J.~Park, Fast point transformer, in: Proceedings of the {IEEE/CVF} Conference on Computer Vision and Pattern Recognition (CVPR), 2022, pp. 16949--16958.

\bibitem{PyramidPC2021}
L.~Hui, H.~Yang, M.~Cheng, J.~Xie, J.~Yang, Pyramid point cloud transformer for large-scale place recognition, 2021 IEEE/CVF International Conference on Computer Vision (ICCV) (2021) 6078--6087.

\bibitem{PointM2AE2022}
R.~Zhang, Z.~Guo, P.~Gao, R.~Fang, B.~Zhao, D.~L. Wang, Y.~J. Qiao, H.~Li, Point-m2ae: Multi-scale masked autoencoders for hierarchical point cloud pre-training, in: Advances in Neural Information Processing Systems, Vol.~35, 2022, pp. 27061--27074.

\bibitem{PointBERT2022}
X.~Yu, L.~Tang, Y.~Rao, T.~Huang, J.~Zhou, J.~Lu, Point-bert: Pre-training 3d point cloud transformers with masked point modeling, in: 2022 IEEE/CVF Conference on Computer Vision and Pattern Recognition (CVPR), 2022, pp. 19291--19300.
\newblock \href {https://doi.org/10.1109/CVPR52688.2022.01871} {\path{doi:10.1109/CVPR52688.2022.01871}}.

\bibitem{Mask2023}
D.~Wang, Z.-X. Yang, Self-supervised point cloud understanding via mask transformer and contrastive learning, IEEE Robotics and Automation Letters 8~(1) (2023) 184--191.
\newblock \href {https://doi.org/10.1109/LRA.2022.3224370} {\path{doi:10.1109/LRA.2022.3224370}}.

\bibitem{2023GTNet}
W.~Zhou, Q.~Wang, W.~Jin, X.~Shi, Y.~He, Gtnet: Graph transformer network for 3d point cloud classification and semantic segmentation (2023).
\newblock \href {http://arxiv.org/abs/2305.15213} {\path{arXiv:2305.15213}}.

\bibitem{2023I2PMAE}
R.~Zhang, L.~Wang, Y.~Qiao, P.~Gao, H.~Li, Learning 3d representations from 2d pre-trained models via image-to-point masked autoencoders, in: Proceedings of the IEEE/CVF Conference on Computer Vision and Pattern Recognition (CVPR), 2023, pp. 21769--21780.

\bibitem{2023AFGCN}
N.~Zhang, Z.~Pan, T.~H. Li, W.~Gao, G.~Li, Improving graph representation for point cloud segmentation via attentive filtering, in: Proceedings of the IEEE/CVF Conference on Computer Vision and Pattern Recognition (CVPR), 2023, pp. 1244--1254.

\bibitem{2023DCNet}
F.~Yin, Z.~Huang, T.~Chen, G.~Luo, G.~Yu, B.~Fu, Dcnet: Large-scale point cloud semantic segmentation with discriminative and efficient feature aggregation, IEEE Transactions on Circuits and Systems for Video Technology 33~(8) (2023) 4083--4095.
\newblock \href {https://doi.org/10.1109/TCSVT.2023.3239541} {\path{doi:10.1109/TCSVT.2023.3239541}}.

\bibitem{2023LCPFormer}
Z.~Huang, Z.~Zhao, B.~Li, J.~Han, Lcpformer: Towards effective 3d point cloud analysis via local context propagation in transformers, IEEE Transactions on Circuits and Systems for Video Technology (2023) 1--13\href {https://doi.org/10.1109/TCSVT.2023.3247506} {\path{doi:10.1109/TCSVT.2023.3247506}}.

\bibitem{Spherical2021}
H.~Lei, N.~Akhtar, A.~Mian, Spherical kernel for efficient graph convolution on 3d point clouds, IEEE Transactions on Pattern Analysis and Machine Intelligence 43~(10) (2021) 3664--3680.
\newblock \href {https://doi.org/10.1109/TPAMI.2020.2983410} {\path{doi:10.1109/TPAMI.2020.2983410}}.

\bibitem{Pointwise2018}
B.-S. Hua, M.-K. Tran, S.-K. Yeung, Pointwise convolutional neural networks, in: 2018 IEEE/CVF Conference on Computer Vision and Pattern Recognition, 2018, pp. 984--993.
\newblock \href {https://doi.org/10.1109/CVPR.2018.00109} {\path{doi:10.1109/CVPR.2018.00109}}.

\bibitem{Mining2018}
Y.~Shen, C.~Feng, Y.~Yang, D.~Tian, Mining point cloud local structures by kernel correlation and graph pooling, in: 2018 IEEE/CVF Conference on Computer Vision and Pattern Recognition, 2018, pp. 4548--4557.
\newblock \href {https://doi.org/10.1109/CVPR.2018.00478} {\path{doi:10.1109/CVPR.2018.00478}}.

\bibitem{Hierarchical2019}
Z.~Liang, M.~Yang, L.~Deng, C.~Wang, B.~Wang, Hierarchical depthwise graph convolutional neural network for 3d semantic segmentation of point clouds, in: 2019 International Conference on Robotics and Automation (ICRA), 2019, pp. 8152--8158.
\newblock \href {https://doi.org/10.1109/ICRA.2019.8794052} {\path{doi:10.1109/ICRA.2019.8794052}}.

\bibitem{Deformable2020}
Z.-H. Lin, S.-Y. Huang, Y.-C.~F. Wang, Convolution in the cloud: Learning deformable kernels in 3d graph convolution networks for point cloud analysis, in: 2020 IEEE/CVF Conference on Computer Vision and Pattern Recognition (CVPR), 2020, pp. 1797--1806.
\newblock \href {https://doi.org/10.1109/CVPR42600.2020.00187} {\path{doi:10.1109/CVPR42600.2020.00187}}.

\bibitem{Point2Node2019}
W.~Han, C.~Wen, C.~Wang, X.~Li, Q.~Li, Point2node: Correlation learning of dynamic-node for point cloud feature modeling, in: AAAI Conference on Artificial Intelligence, 2019.

\bibitem{Spectral2018}
C.~Wang, B.~Samari, K.~Siddiqi, Local spectral graph convolution for point set feature learning, in: European Conference on Computer Vision, 2018.

\bibitem{Oversegmentation2019}
L.~Landrieu, M.~Boussaha, Point cloud oversegmentation with graph-structured deep metric learning, in: 2019 IEEE/CVF Conference on Computer Vision and Pattern Recognition (CVPR), 2019, pp. 7432--7441.
\newblock \href {https://doi.org/10.1109/CVPR.2019.00762} {\path{doi:10.1109/CVPR.2019.00762}}.

\bibitem{Pytorch2019}
A.~Paszke, S.~Gross, F.~Massa, A.~Lerer, J.~Bradbury, G.~Chanan, T.~Killeen, Z.~Lin, N.~Gimelshein, L.~Antiga, A.~Desmaison, A.~Kopf, E.~Yang, Z.~DeVito, M.~Raison, A.~Tejani, S.~Chilamkurthy, B.~Steiner, L.~Fang, J.~Bai, S.~Chintala, Pytorch: An imperative style, high-performance deep learning library, in: Advances in Neural Information Processing Systems, Vol.~32, 2019.

\bibitem{PCNN2018}
M.~Atzmon, H.~Maron, Y.~Lipman, Point convolutional neural networks by extension operators, ACM Trans. Graph. 37~(4) (jul 2018).
\newblock \href {https://doi.org/10.1145/3197517.3201301} {\path{doi:10.1145/3197517.3201301}}.

\bibitem{2022CSANet}
G.~Wang, Q.~Zhai, H.~Liu, \href{https://www.sciencedirect.com/science/article/pii/S0950705122003616}{Cross self-attention network for 3d point cloud}, Knowledge-Based Systems 247 (2022) 108769.
\newblock \href {https://doi.org/https://doi.org/10.1016/j.knosys.2022.108769} {\path{doi:https://doi.org/10.1016/j.knosys.2022.108769}}.
\newline\urlprefix\url{https://www.sciencedirect.com/science/article/pii/S0950705122003616}

\bibitem{RSCNN2019}
Y.~Liu, B.~Fan, S.~Xiang, C.~Pan, Relation-shape convolutional neural network for point cloud analysis, in: 2019 IEEE/CVF Conference on Computer Vision and Pattern Recognition (CVPR), 2019, pp. 8887--8896.
\newblock \href {https://doi.org/10.1109/CVPR.2019.00910} {\path{doi:10.1109/CVPR.2019.00910}}.

\bibitem{PCT2021}
M.-H. Guo, J.-X. Cai, Z.-N. Liu, T.-J. Mu, R.~R. Martin, S.-M. Hu, Pct: Point cloud transformer, Computational Visual Media 7~(2) (2021) 187--199.
\newblock \href {https://doi.org/10.1007/s41095-021-0229-5} {\path{doi:10.1007/s41095-021-0229-5}}.

\bibitem{2022DTONet}
R.~Hu, B.~Yang, H.~Ye, F.~Cao, C.~Wen, Q.~Zhang, \href{https://www.sciencedirect.com/science/article/pii/S0950705122004245}{Decouple the object: Component-level semantic recognizer for point clouds classification}, Knowledge-Based Systems 248 (2022) 108887.
\newblock \href {https://doi.org/https://doi.org/10.1016/j.knosys.2022.108887} {\path{doi:https://doi.org/10.1016/j.knosys.2022.108887}}.
\newline\urlprefix\url{https://www.sciencedirect.com/science/article/pii/S0950705122004245}

\bibitem{SO-Net2018}
J.~Li, B.~M. Chen, G.~H. Lee, So-net: Self-organizing network for point cloud analysis, in: 2018 IEEE/CVF Conference on Computer Vision and Pattern Recognition, 2018, pp. 9397--9406.
\newblock \href {https://doi.org/10.1109/CVPR.2018.00979} {\path{doi:10.1109/CVPR.2018.00979}}.

\bibitem{2024PointConT}
Y.~Liu, B.~Tian, Y.~Lv, L.~Li, F.-Y. Wang, Point cloud classification using content-based transformer via clustering in feature space, IEEE/CAA Journal of Automatica Sinica 11~(1) (2024) 231--239.
\newblock \href {https://doi.org/10.1109/JAS.2023.123432} {\path{doi:10.1109/JAS.2023.123432}}.

\bibitem{pointmixer2021}
J.~Choe, C.~Park, F.~Rameau, J.~Park, I.~S. Kweon, Pointmixer: Mlp-mixer for point cloud understanding, arXiv preprint arXiv:2111.11187 (2021).

\bibitem{point-trans2021}
H.~Zhao, L.~Jiang, J.~Jia, P.~Torr, V.~Koltun, Point transformer, in: 2021 IEEE/CVF International Conference on Computer Vision (ICCV), 2021, pp. 16239--16248.
\newblock \href {https://doi.org/10.1109/ICCV48922.2021.01595} {\path{doi:10.1109/ICCV48922.2021.01595}}.

\bibitem{CurveNet2021}
T.~Xiang, C.~Zhang, Y.~Song, J.~Yu, W.~Cai, Walk in the cloud: Learning curves for point clouds shape analysis, in: Proceedings of the IEEE/CVF International Conference on Computer Vision (ICCV), 2021, pp. 915--924.

\bibitem{PointASNL2020}
X.~Yan, C.~Zheng, Z.~Li, S.~Wang, S.~Cui, Pointasnl: Robust point clouds processing using nonlocal neural networks with adaptive sampling, in: Proceedings of the IEEE/CVF Conference on Computer Vision and Pattern Recognition (CVPR), 2020.

\bibitem{SEGCloud2017}
L.~P. Tchapmi, C.~B. Choy, I.~Armeni, J.~Gwak, S.~Savarese, Segcloud: Semantic segmentation of 3d point clouds, 2017 International Conference on 3D Vision (3DV) (2017) 537--547.

\bibitem{PointWeb2019}
H.~Zhao, L.~Jiang, C.-W. Fu, J.~Jia, Pointweb: Enhancing local neighborhood features for point cloud processing, in: 2019 IEEE/CVF Conference on Computer Vision and Pattern Recognition (CVPR), 2019, pp. 5560--5568.

\bibitem{HPEIN2019}
L.~Jiang, H.~Zhao, S.~Liu, X.~Shen, C.-W. Fu, J.~Jia, Hierarchical point-edge interaction network for point cloud semantic segmentation, 2019 IEEE/CVF International Conference on Computer Vision (ICCV) (2019) 10432--10440.

\bibitem{Graph2019}
L.~Wang, Y.~Huang, Y.~Hou, S.~Zhang, J.~Shan, Graph attention convolution for point cloud semantic segmentation, in: 2019 IEEE/CVF Conference on Computer Vision and Pattern Recognition (CVPR), 2019, pp. 10288--10297.
\newblock \href {https://doi.org/10.1109/CVPR.2019.01054} {\path{doi:10.1109/CVPR.2019.01054}}.

\bibitem{GAPointNet2021}
C.~Chen, L.~Z. Fragonara, A.~Tsourdos, Gapointnet: Graph attention based point neural network for exploiting local feature of point cloud, Neurocomputing 438 (2021) 122--132.
\newblock \href {https://doi.org/https://doi.org/10.1016/j.neucom.2021.01.095} {\path{doi:https://doi.org/10.1016/j.neucom.2021.01.095}}.

\bibitem{GIN2019}
X.~Keyulu, H.~Weihua, L.~Jure, J.~Stefanie, How powerful are graph neural networks?, in: International Conference on Learning Representations, 2019.

\bibitem{SAGE2017}
W.~L. Hamilton, R.~Ying, J.~Leskovec, Inductive representation learning on large graphs, in: Proceedings of the 31st International Conference on Neural Information Processing Systems, NIPS'17, Red Hook, NY, USA, 2017, p. 1025–1035.

\end{thebibliography}
\end{document}